\documentclass{article}

\usepackage{arxiv}
\usepackage{bm}
\usepackage{listings}
\usepackage{float}
\usepackage{amsmath}
\usepackage{amssymb}
\usepackage{caption} 
\usepackage{tabularx}
\usepackage{booktabs}
\usepackage{ragged2e}
\usepackage{array}
\usepackage{chngcntr}
\usepackage{makecell}
\usepackage{subcaption}
\usepackage{natbib}

\usepackage{svg}

\usepackage{graphicx}
\usepackage[utf8]{inputenc} % allow utf-8 input
\usepackage[T1]{fontenc}    % use 8-bit T1 fonts

\graphicspath{ {images/} }
\usepackage{lineno}

\usepackage{hyperref}
\usepackage{url}
\usepackage[utf8]{inputenc} % allow utf-8 input
\usepackage[T1]{fontenc}    % use 8-bit T1 fonts
\usepackage{hyperref}       % hyperlinks
\usepackage{url}            % simple URL typesetting
\usepackage{booktabs}       % professional-quality tables
\usepackage{amsfonts}       % blackboard math symbols
\usepackage{nicefrac}       % compact symbols for 1/2, etc.
\usepackage{microtype}      % microtypography
\usepackage{xcolor}         % colors
\usepackage{amsmath}
\usepackage{amssymb}
\usepackage{amsthm}
\usepackage{graphicx}
\usepackage{subcaption}
\usepackage{colortbl}
\usepackage{algorithm}
\usepackage{subcaption}
\usepackage{tabularx}
\usepackage{amsmath,amssymb,amsthm,bbm}
\usepackage{multirow}
\usepackage{adjustbox}
\usepackage{algorithm}
\usepackage{algpseudocode}
\usepackage{amsmath}
\usepackage[noEnd=false]{algpseudocodex}
\usepackage[toc,page]{appendix} % For appendix and TOC in the appendix
\usepackage{tocloft} % For customizing TOC
\usepackage{titlesec}
\usepackage{booktabs}
\usepackage{threeparttable}
\usepackage{enumitem}
\setlist[enumerate]{left=10pt}
\usepackage[font=small]{caption}
\usepackage{wrapfig}
\usepackage{makecell}
\usepackage{tocloft}
\usepackage{xcolor}
\usepackage{setspace}

\makeatletter
\newcommand\midfootnotesize{%
  \@setfontsize\midfootnotesize{8}{9}%
}
\makeatother

\definecolor{codegreen}{rgb}{0,0.6,0}
\definecolor{codegray}{rgb}{0.5,0.5,0.5}
\definecolor{codepurple}{rgb}{0.58,0,0.82}
\definecolor{backcolour}{rgb}{0.95,0.95,0.92}

\lstdefinestyle{mystyle}{
    backgroundcolor=\color{backcolour},   
    commentstyle=\color{codegreen},
    keywordstyle=\color{magenta},
    numberstyle=\tiny\color{codegray},
    stringstyle=\color{codepurple},
    basicstyle=\ttfamily\footnotesize,
    breakatwhitespace=false,         
    breaklines=true,                 
    captionpos=b,                    
    keepspaces=true,                 
    numbers=left,                    
    numbersep=5pt,                  
    showspaces=false,                
    showstringspaces=false,
    showtabs=false,                  
    tabsize=2
}

\title{AReUReDi: Annealed Rectified Updates for Refining Discrete Flows with Multi-Objective Guidance} 

\author
{Tong Chen,$^{1}$ Yinuo Zhang,$^{2}$ \textbf{Pranam Chatterjee}$^{1,3,\dag}$\\
\\$^{1}$Department of Computer and Information Science, University of Pennsylvania
\\$^{2}$Centre for Computational Biology, Duke-NUS Medical School, Singapore
\\$^{3}$Department of Bioengineering, University of Pennsylvania
\\\\\normalsize{$^\dag$Corresponding author: pranam@seas.upenn.edu}
}

\begin{document} 


% \baselineskip24pt


\maketitle

\begin{abstract}
Designing sequences that satisfy multiple, often conflicting, objectives is a central challenge in therapeutic and biomolecular engineering. Existing generative frameworks largely operate in continuous spaces with single-objective guidance, while discrete approaches lack guarantees for multi-objective Pareto optimality. We introduce \textbf{AReUReDi} (\textbf{A}nnealed \textbf{Re}ctified \textbf{U}pdates for \textbf{Re}fining \textbf{Di}screte Flows), a discrete optimization algorithm with theoretical guarantees of convergence to the Pareto front. Building on Rectified Discrete Flows (ReDi), AReUReDi combines Tchebycheff scalarization, locally balanced proposals, and annealed Metropolis-Hastings updates to bias sampling toward Pareto-optimal states while preserving distributional invariance. Applied to peptide and SMILES sequence design, AReUReDi simultaneously optimizes up to five therapeutic properties (including affinity, solubility, hemolysis, half-life, and non-fouling) and outperforms both evolutionary and diffusion-based baselines. These results establish AReUReDi as a powerful, sequence-based framework for multi-property biomolecule generation.
\end{abstract}

\section{Introduction}
The design of biological sequences must account for multiple, often conflicting, objectives \citep{naseri2020application}. Therapeutic peptides, for example, must combine high binding affinity with low toxicity and favorable pharmacokinetics \citep{tominaga2024designing, tang2025peptune}; CRISPR guide RNAs require both high on-target activity and minimal off-target effects \citep{mohr2016crispr, schmidt2025genome}; and synthetic promoters must deliver strong expression while remaining tissue-specific \citep{artemyev2024synthetic}. These examples illustrate that biomolecular engineering is inherently a multi-objective optimization problem.

Yet, most computational frameworks continue to optimize single objectives in isolation \citep{zhou2019all, nehdi2020novel, nisonoff2025guidance}. While such approaches can reduce toxicity \citep{kreiser2020therapeutic, sharma2022toxinpred2} or improve thermostability \citep{komp2025neural}, they often create adverse trade-offs: high-affinity peptides may be insoluble or hemolytic, and stabilized proteins may lose specificity \citep{bigi2023toxicity, rinauro2024misfolded}. Black-box multi-objective optimization (MOO) methods such as evolutionary search and Bayesian optimization have long been applied to molecular design \citep{zitzler1998multiobjective, deb2011multi, ueno2016combo, frisby2021bayesian}, but these approaches scale poorly in high-dimensional sequence spaces.

To overcome this, recent generative approaches have incorporated controllable multi-objective sampling \citep{li2018multi, sousa2021combining, yao2024proud}. For instance, ParetoFlow \citep{yuan2024paretoflow} leverages continuous-space flow matching to generate Pareto-optimal samples. However, extending such guarantees to biological sequences is challenging, since discrete data typically require embedding into continuous manifolds, which distorts token-level structure and complicates property-based guidance \citep{beliakov2007challenges, michael2024continuous}.

A more direct path lies in discrete flow models \citep{campbell2024generative, gat2024discrete, dunn2024exploring}. These models define probability paths over categorical state spaces, either through simplex-based interpolations \citep{stark2024dirichlet, davis2024fisher, tang2025gumbel} or jump-process flows that learn token-level transition rates \citep{campbell2024generative, gat2024discrete}. Recent advances have shown their promise for controllable single-objective generation \citep{nisonoff2025guidance, tang2025gumbel}, but no framework yet achieves Pareto guidance across multiple objectives.

Here, the notion of rectification provides a crucial building block. In the continuous setting, \textit{Rectified Flows} \citep{liu2023flow} learn to straighten ODE paths between distributions, thereby reducing convex transport costs and enabling efficient few-step or even one-step sampling. Recently, \textbf{ReDi} (\textit{Rectified Discrete Flows}) \citep{yoo2025redi} extended this principle to discrete domains. By iteratively refining the coupling between source and target distributions, ReDi provably reduces factorization error (quantified as conditional total correlation) while maintaining distributional fidelity. This makes ReDi highly effective for efficient discrete sequence generation. However, ReDi does not address the multi-objective setting, as it lacks a mechanism to steer sampling toward the \textit{Pareto front}, where improvements in one objective cannot be made without degrading another. This is a critical limitation for biomolecular design, where trade-offs define practical success.

To address this, we introduce \textbf{AReUReDi} (\textbf{A}nnealed \textbf{Re}ctified \textbf{U}pdates for \textbf{Re}fining \textbf{Di}screte Flows), a new framework that extends rectified discrete flows with multi-objective guidance. AReUReDi integrates three innovations: (i) \textit{annealed Tchebycheff scalarization}, which gradually sharpens the focus on balanced solutions across objectives \citep{lin2024few}; (ii) \textit{locally balanced proposals}, which combine the generative prior of ReDi with multi-objective guidance while ensuring reversibility; and (iii) \textit{Metropolis-Hastings updates}, which preserve exact distributional invariance and guarantee convergence to Pareto-optimal states. Together, these mechanisms refine rectified discrete flows into a principled Pareto sampler.

\noindent Our key contributions are:
\begin{enumerate}
\item We propose AReUReDi, the first multi-objective extension of rectified discrete flows, integrating annealed scalarization, locally balanced proposals, and MCMC updates.
\item We provide theoretical guarantees that AReUReDi preserves distributional invariance and converges to the Pareto front with full coverage.
\item We demonstrate that AReUReDi can optimize up to five competing biological properties simultaneously, including affinity, solubility, hemolysis, half-life, and non-fouling, for therapeutic peptide design.
\item We benchmark AReUReDi against classical MOO algorithms and state-of-the-art discrete diffusion approaches, showing superior trade-off navigation and biologically plausible peptide sequence designs.
\end{enumerate}

\section{Preliminaries}

\subsection{Discrete Flow Matching}
Let $\mathcal{S}=V^L$ denote the discrete state space, where $V$ is a vocabulary of size $K$ and each $x=(x_1,\dots,x_L)\in\mathcal{S}$ is a sequence of tokens. 
A \textit{discrete flow matching (DFM)} model \citep{campbell2024generative, gat2024discrete, dunn2024exploring} defines a probability path $\{p_t\}_{t\in[0,1]}$ interpolating between a simple source distribution $p_0$ and a target distribution $p_1$ by means of a coupling $\pi(x_0,x_1)$ and conditional bridge distributions $p_t(x_t\mid x_0,x_1)$. 
The model is trained to approximate conditional transitions $p_{s|t}(x_s\mid x_t)$ for $0\le t<s\le 1$.  

Since the joint distribution over $L$ coordinates is intractable, DFMs employ a factorization
\[
p_{s|t}(x_s\mid x_t)\;\approx\;\prod_{i=1}^L p_{s|t}\!\left(x_s^i \mid x_t \right),
\]
which introduces a discrepancy measured by the conditional total correlation
\[
\mathrm{TC}_{s|t} \;=\; \mathrm{KL}\!\left(
p_{s|t}(x_s\mid x_t)\;\middle\|\;\prod_{i=1}^L p_{s|t}(x_s^i\mid x_t)
\right).
\]
This quantity captures the inter-dimensional dependencies neglected under factorization, and grows with larger step sizes \citep{stark2024dirichlet, davis2024fisher, tang2025gumbel}. As a result, DFMs are accurate in the many-step regime but degrade under few-step or one-step generation.

\subsection{Rectified Discrete Flow}
To mitigate factorization error, \textbf{Rectified Discrete Flow (ReDi)} \citep{yoo2025redi} introduces an iterative rectification of the coupling $\pi$. 
Starting from an initial coupling $\pi^{(0)}(x_0,x_1)$, a DFM is trained under $\pi^{(k)}$ to produce new source--target pairs, defining an empirical joint distribution $\hat{\pi}^{(k)}$. 
The coupling is then updated via
\[
\pi^{(k+1)}(x_0,x_1)
\;\propto\;
\pi^{(k)}(x_0,x_1)\,
\frac{p_{\theta^{(k)}}(x_1\mid x_0)}{p_{\theta^{(k)}}(x_1)},
\]
where $p_{\theta^{(k)}}(x_1\mid x_0)$ is the conditional distribution learned at iteration $k$. 
This yields a sequence of couplings $\{\pi^{(k)}\}_{k\ge0}$ with provably decreasing conditional TC,
\[
\mathrm{TC}_{s|t}(\pi^{(k+1)}) \;\le\; \mathrm{TC}_{s|t}(\pi^{(k)}).
\]
By progressively reducing factorization error, ReDi produces a well-calibrated base distribution $p_1$ with low inter-dimensional correlation. This base distribution provides reliable marginal transition probabilities $p_t^i(\cdot\mid x_t)$ for each coordinate $i$ at time $t$, which serve as the generative prior in the AReUReDi framework. 
Rectification follows the same principle as \textit{Rectified Flow} in continuous domains \citep{liu2023flow}, where iterative refinement straightens ODE paths and decreases transport costs.

\subsection{Multi-Objective Setup}
In biomolecular design and related applications, the generation task is inherently multi-objective \citep{zitzler1998multiobjective, deb2011multi, frisby2021bayesian}. 
Let $s_1,\dots,s_N:\mathcal{S}\to\mathbb{R}$ denote $N$ scalar objectives, and let $\tilde s_n(x)$ be their normalized counterparts mapping into $[0,1]$ to ensure comparability. 
Given weights $\omega\in\Delta^{N-1}$, the Tchebycheff scalarization is defined as
\[
S_\omega(x) \;=\; \min_{1\le n\le N} \;\omega_n\, \tilde s_n(x),
\]
which balances objectives by rewarding solutions that are simultaneously strong across all metrics rather than excelling in just one \citep{Miettinen1999}. 
This scalarization will serve as the core of the guidance mechanism in AReUReDi.

\begin{figure}[t]
    \centering
    \includegraphics[width=\textwidth]{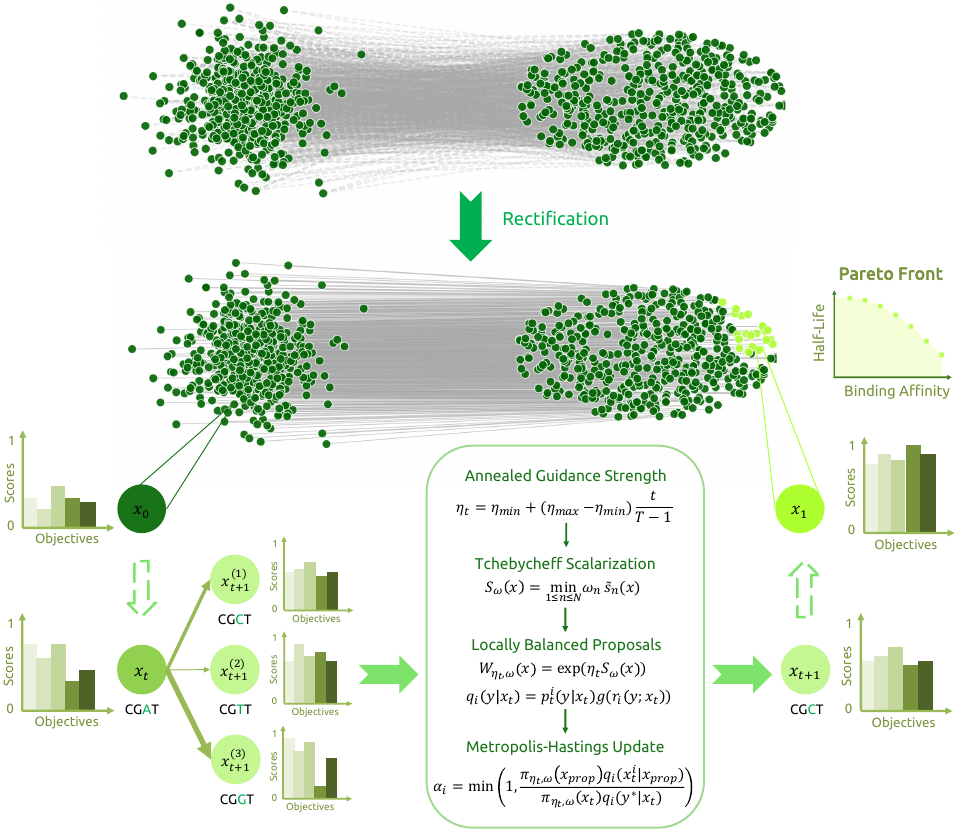}
    \caption{\textbf{AReUReDi.} Discrete flow matching is first rectified to reduce conditional total correlation. At each timestep, candidate single-position mutations with ReDi-predicted probabilities (visualized by arrows of varying thickness) are evaluated by multiple objective functions. A locally balanced proposal is then constructed using Tchebycheff scalarization with annealed guidance strength, and the next state is selected via a Metropolis-Hastings update. This iterative process drives the generated sequences toward the Pareto front.}
    \label{fig:areuredi}
    \vskip -0.1in
\end{figure}

\section{AReUReDi: Annealed Rectified Updates for Refining Discrete Flows}
With an efficient discrete flow-based generation framework in hand, we develop AReUReDi that extends ReDi \citep{yoo2025redi} to the multi-objective optimization setting, where the goal is to generate discrete samples that approximate the Pareto front of multiple competing objectives. Starting from a pre-trained ReDi model, AReUReDi incorporates annealed guidance, locally balanced proposals, and Metropolis-Hastings updates to progressively bias the sampling process toward Pareto-optimal states while preserving the probabilistic guarantees of the underlying flow (Figure \ref{fig:areuredi}, Algorithm \ref{algo:areuredi}).

\subsection{Problem Setup}
Let the discrete search space be $\mathcal{S}=\mathcal{V}^L$, where $\mathcal{V}$ is a finite vocabulary of size $K$ and each state $x=(x_1,\dots,x_L)\in \mathcal{S}$ is a sequence of tokens. We assume access to a pre-trained ReDi model that provides marginal transition probabilities $p_t^i(\cdot\mid x_t)$ for each position $i$ and time $t$. In addition, we are given $N$ pre-trained scalar objective functions $s_n: \mathcal{S} \to \mathbb{R}$, where $n=1,\dots,N$, and $\tilde s_n(x)$ are their normalized counterparts with outputs mapped to $[0,1]$ to support balanced updates for each objective. The sampling task is to construct a Markov chain whose stationary distribution concentrates on states that approximate the Pareto front of the normalized objectives $\tilde s_1,\dots,\tilde s_N$.

\subsection{Annealed Multi-Objective Guidance}
To direct sampling toward the Pareto front, AReUReDi introduces a scalarized reward
\[
S_{\omega}(x)=\min_{1\le n\le N} \omega_n\, \tilde s_n(x),
\]
where the weight vector $\omega = [\omega_1, \dots,\omega_N]$ lies in the probability simplex $\Delta^{N-1}$ and balances the different objectives. This Tchebycheff scalarization promotes solutions that are simultaneously strong across all objectives rather than excelling in only a subset \citep{Miettinen1999}. The scalarized reward is converted into a guidance weight
\[
W_{\eta_t,\omega}(x)=\exp\big(\eta_t S_{\omega}(x)\big),
\]
where the parameter $\eta_t>0$ controls the strength of the guidance at each iteration $t$. AReUReDi incorporates an annealing schedule for $\eta_t$: 
\[
\eta_t \;=\; \eta_{\min}+\bigl(\eta_{\max}-\eta_{\min}\bigr)\frac{t}{\,T-1\,},
\]
so that the chain begins with a small value of $\eta_t$ to encourage wide exploration of the state space and gradually increases $\eta_t$ to focus sampling on high-quality Pareto candidates. This annealing strategy mirrors simulated annealing but operates directly on the scalarized objectives within the discrete flow framework.

\subsection{Locally Balanced Proposals}
Given the current state $x_t$, AReUReDi updates one coordinate $i\in\{1,\dots,L\}$ at a time using a locally balanced proposal that blends the generative prior of ReDi with the multi-objective guidance. First, a candidate set of replacement tokens is drawn from the ReDi marginal $p_t^i(\cdot\mid x_t)$, optionally pruned using top-$\text{p}$ to retain only the most promising alternatives for computational efficiency. For each candidate token $y$, the algorithm computes the ratio
\[
r_i(y;x_t) =
\frac{W_{\eta_t,\omega}\big(x_t^{(i\leftarrow y)}\big)}
     {W_{\eta_t,\omega}(x_t)},
\]
which measures the change in scalarized reward if $x_t^i$ were replaced by $y$. The ratio $r_i(y;x_t)$ is then transformed by a balancing function $g:\mathbb{R}_+\to\mathbb{R}_+$ that satisfies the symmetry condition $g(u)=u\,g(1/u)$. Typical choices include Barker’s function $g(u)=\frac{u}{1+u}$ and the square-root function $g(u)=\sqrt{u}$. This symmetry ensures that the resulting Markov chain admits the desired stationary distribution. Using the balanced function, the unnormalized proposal for a candidate token $y$ takes the form
\[
\tilde q_i(y\mid x_t)= p_t^i(y\mid x_t)\,g\big(r_i(y;x_t)\big),
\]
which is then normalized over the candidate set to yield the final proposal distribution $q_i(y\mid x_t)$. This construction allows the proposal to favor states with higher scalarized reward while remaining reversible with respect to the target distribution.

\subsection{Metropolis-Hastings Update}
A candidate token $y^\star$ is drawn from the final proposal distribution $q_i(\cdot\mid x_t)$ and forms the proposed state $x_{\mathrm{prop}} = x_t^{(i\leftarrow y^\star)}$. The proposal is accepted with the standard Metropolis-Hastings probability \citep{hastings1970monte}
\[
\alpha_i(x_t,x_{\mathrm{prop}})=
\min\left\{1,\,
\frac{\pi_{\eta_t,\omega}(x_{\mathrm{prop}})\,
      q_i(x_t^i\mid x_{\mathrm{prop}})}
     {\pi_{\eta_t,\omega}(x_t)\,
      q_i(y^\star\mid x_t)}
\right\},
\]
where we define $ \pi_{\eta_t,\omega}(x) \;\propto\; p_1(x)\,W_{\eta_t,\omega}(x) \;=\; p_1(x)\,\exp\!\big(\eta_t S_{\omega}(x)\big)$. With Barker’s balancing function, the acceptance probability simplifies to one, ensuring automatic acceptance of proposals and faster mixing. Other choices, such as the square-root function, trade higher acceptance rates for more conservative moves.

The annealed, locally balanced updates are repeated for $T$ iterations and end with the final sample $x_1$ whose objective scores are jointly optimized. Building on the ReDi model’s well-calibrated base distribution with low inter-dimensional correlation, AReUReDi safely biases this base toward Pareto-optimal regions while preserving full coverage of the state space, thereby guaranteeing convergence to Pareto-optimal solutions with complete coverage of the Pareto front (Proof \ref{sec:proofs-main}). 

\section{Experiments}
To the best of our knowledge, no public datasets exist for benchmarking multi-objective optimization algorithms on biological sequences. We therefore developed two benchmarks to evaluate AReUReDi, focusing on the generation of wild-type peptide sequences and chemically-modified peptide SMILES. These tasks are supported by two core components: the generative models described in Section \ref{pepredi_SMILESReDi} and the objective-scoring models validated in Section \ref{sec:score_models}. Leveraging these models, we demonstrate AReUReDi’s efficacy on a wide range of tasks and examples.

Although AReUReDi provides theoretical guarantees of Pareto optimality and full coverage, in practice, these guarantees hold only in the limit of an infinitely long Markov chain. Reaching the Pareto front with high probability can therefore require a vast number of sampling steps. To improve sampling efficiency in all reported experiments, we introduce a monotonicity constraint that accepts only token updates that increase the weighted sum of the current objective scores. Empirical results prove the accelerated convergence toward high-quality Pareto solutions without altering the underlying optimization objectives (Table \ref{tab:ablation_constraint}). Therefore, this monotonicity constraint was involved in all the following experiments. 

\subsection{PepReDi and SMILESReDi Generate Diverse and Biologically Plausible Sequences}
\label{pepredi_SMILESReDi}

To enable the efficient generation of peptide binders, we developed an unconditional peptide generator, \textbf{PepReDi}, based on the ReDi framework. The model backbone of PepReDi is a Diffusion Transformer (DiT) architecture \citep{Peebles2022DiT}. We trained PepReDi on a custom dataset comprising approximately 15,000 peptides from the PepNN and BioLip2 datasets, as well as sequences from the PPIRef dataset, with lengths ranging from 6 to 49 amino acids \citep{abdin2022pepnn, zhang2024biolip2, bushuiev2023learning}. Using this trained model, we generated new data couplings containing 10,000 sequences for each peptide length and used them to fine-tune PepReDi in an iterative rectification procedure. This rectification was performed three times and yielded substantial improvements in training loss, validation negative log-likelihood (NLL), perplexity (PPL), and conditional TC (Table \ref{tab:pepredi}). Notably, the conditional TC rises after the first rectification, likely due to the distributional shift from the large, model-generated coupling, whose absolute TC can be higher even though ReDi guarantees a monotonic decrease within each coupling. The low validation NLL and PPL metrics showcase PepReDi's reliability to generate biologically plausible wild-type peptide sequences. 

\begin{table}[t]
\small
\centering
\caption{Training and validation performance of PepReDi over successive rectification rounds. Each row reports the training loss, validation negative log-likelihood (NLL), validation perplexity (PPL), and conditional total correlation (TC). PepReDi without superscript denotes the base model, while PepReDi$^1$, PepReDi$^3$, PepReDi$^3$ indicate the first, second, and third rounds of rectification, respectively.}
\label{tab:pepredi}
\begin{tabular}{lcccc}
\toprule
& \textbf{Train Loss} & \textbf{Val NLL} & \textbf{Val PPL} & \textbf{Conditional TC} \\
\midrule
PepReDi& 1.6567 & 1.6458 & 5.19 & 10.6027 \\ \midrule
PepReDi$^1$& 1.6170 & 1.6101 & 5.00 & 12.6250 \\
PepReDi$^2$& 1.5347 & 1.5238 & 4.59 & 11.7279 \\
PepReDi$^3$& \textbf{1.3538} & \textbf{1.3548} & \textbf{3.88} & \textbf{11.2339} \\
\bottomrule
\end{tabular}
\end{table}
\renewcommand{\arraystretch}{1.3}
\begin{table}[t]
\small
\centering
\caption{Evaluation metrics for the generative quality of peptide SMILES sequences of max token length set to 200. SMILESReDi without superscription denotes the base model, while SMILESReDi$^1$ refers to the model that has undergone one round of rectification.}
\label{tab:gen_quality}
\begin{tabular}{lcccc}
\toprule
\textbf{Model} & \textbf{Validity (↑)} & \textbf{Uniqueness (↑)} & \textbf{Diversity (↑)} & \textbf{SNN (↓)} \\
\midrule
Data     & 1.000 & 1.000 & 0.885 & 1.000 \\
PepMDLM  & 0.450 & 1.000 & 0.705 & 0.513 \\
\textbf{SMILESReDi}& \textbf{0.763} & 1.000 & 0.719& 0.593\\ 
\textbf{SMILESReDi$^1$}& \textbf{0.986}& 1.000 & 0.665& 0.579\\ \midrule
PepTune  & 1.000 & 1.000 & 0.677 & 0.486 \\
 \textbf{AReUReDi}& 1.000 & 1.000 & \textbf{0.789}&\textbf{0.392}\\ 
 \bottomrule
\end{tabular}
\label{tab:SMILESReDi}
\end{table}

SMILESReDi adopts the same backbone structure as PepReDi, enhanced with Rotary Positional Embeddings (RoPE), which effectively captures the relative inter-token interactions in peptide SMILES \citep{su2024roformer}. SMILESReDi also incorporates a time-dependent noising schedule to improve its capability to generate valid peptide SMILES sequences (Section \ref{sec:SMILESReDi}). We applied the same training data as PepMDLM, a state-of-the-art diffusion model that generates valid peptide SMILES sequences \citep{tang2025peptune}. After only two training epochs, SMILESReDi converged to a validation NLL of 0.722 and achieved a sampling validity of 76.3\% using just 16 generation steps. One hundred SMILES sequences were then generated by the trained SMILESReDi for each length from 4 to 1035, forming a large and diverse new data coupling. Following a single round of rectification, the validation NLL further decreased to 0.608, and the sampling validity rose dramatically to 98.6\% with 16 steps and 100\% with 32 steps (Table \ref{tab:SMILESReDi}). While its similarity-to-nearest-neighbor (SNN) score and diversity are comparable to those of PepMDLM (details on metrics are provided in Section \ref{sec:SMILESReDi}), SMILESReDi substantially outperforms PepMDLM in validity, highlighting its superior capability of generating diverse chemically-modified peptide SMILES sequences. 

\subsection{AReUReDi effectively balances each objective trade-off}
With pre-trained PepReDi in hand, we first focus on validating AReUReDi's capability of balancing multiple conflicting objectives. We performed two sets of experiments for wild-type peptide binder generation with three property guidance, and in ablation experiment settings, we removed one or more objectives. In the binder design task for target 7LUL (hemolysis, solubility, affinity guidance; Table~\ref{tab:ablation_7LUL}), omitting any single guidance causes a collapse in that property, while the remaining guided metrics may modestly improve. Likewise, in the binder design task for target CLK1 (affinity, non‑fouling, half‑life guidance; Table~\ref{tab:ablation_CLK1}), disabling non‑fouling guidance allows half‑life to exceed 96 hours but drives non‑fouling near zero, and disabling half‑life guidance preserves non‑fouling yet reduces half‑life below 2 hours. In contrast, enabling all guidance signals produces the most balanced profiles across all objectives. These results confirm that AReUReDi precisely targets chosen objectives while preserving the flexibility to navigate conflicting objectives and push samples toward the Pareto front.

\begin{figure}[t]
    \centering
    \includegraphics[width=\textwidth]{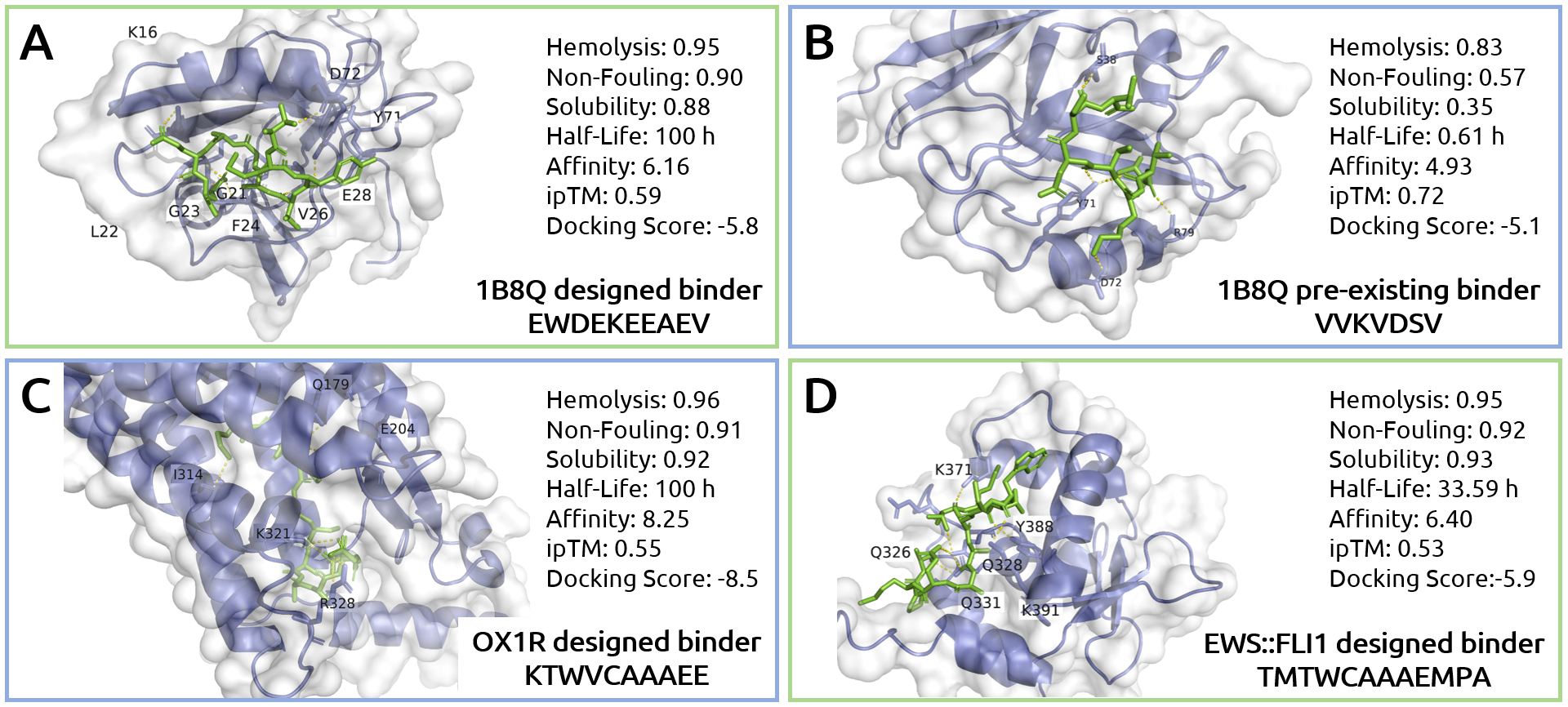}
    \vspace{0.3em}
      \includegraphics[width=\textwidth]{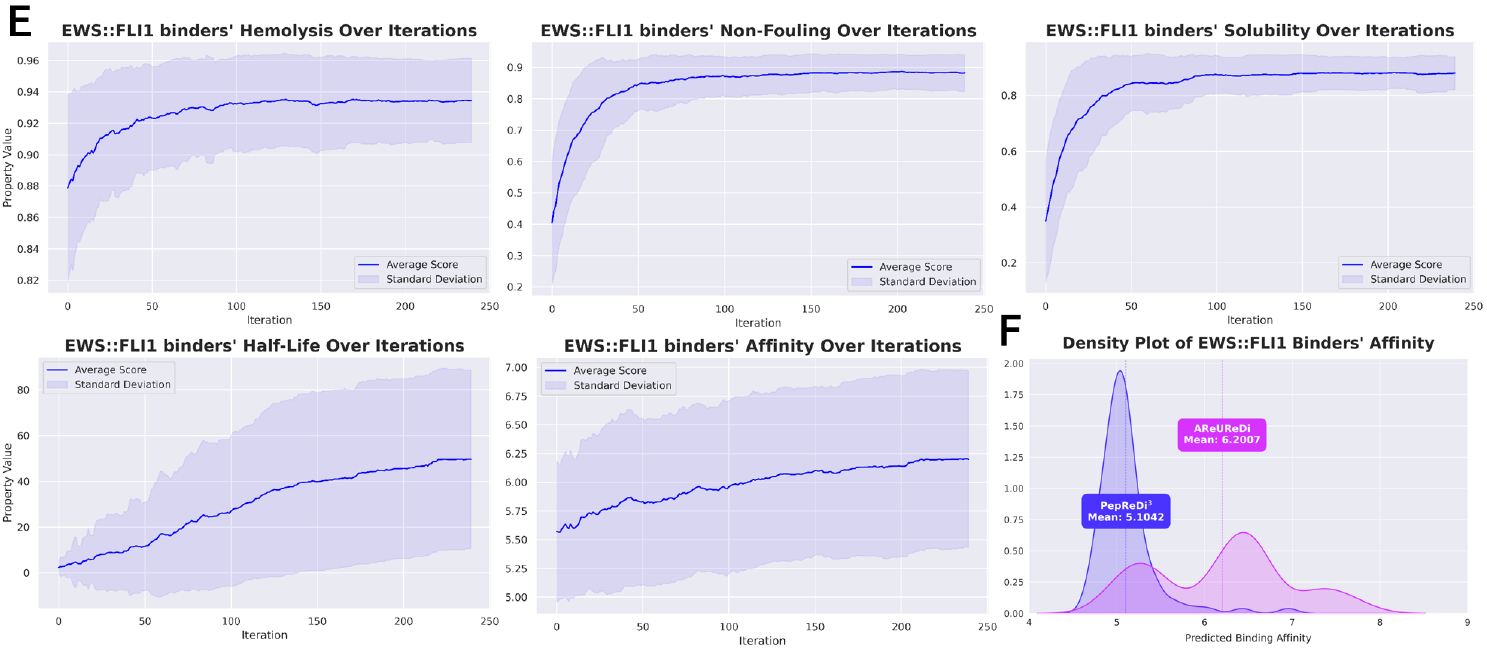}
    \caption{\textbf{(A), (B)} Complex structures of PDB 1B8Q with an AReUReDi-designed binder and its pre-existing binder. \textbf{(C), (D)} Complex structures of OX1R and EWS::FLI1 with an AReUReDi-designed binder. Five property scores are shown for each binder, along with the ipTM score from AlphaFold3 and docking score from AutoDock VINA. Interacting residues on the target are visualized. \textbf{(E)} Plots showing the mean scores for each property across the number of iterations during AReUReDi's design of binders of length 12-aa for EWS::FLI1. \textbf{(F)} A density plot illustrating the distribution of predicted property scores for AReUReDi-designed EWS::FLI1 binders of length 12-aa, compared to the peptides generated unconditionally by PepReDi$^3$.}
    \label{fig:visualization}
    \vskip -0.1in
\end{figure}

\renewcommand{\arraystretch}{1.3}
\begin{table}[t]
\midfootnotesize
\centering
\caption{AReUReDi generates wild-type peptide binders for 8 diverse protein targets, optimizing five therapeutic properties: hemolysis, non-fouling, solubility, half-life (in hours), and binding affinity. Each value represents the average of 100 AReUReDi-designed binders.}
\label{tab:binder_properties}
\begin{tabular}{ccccccc}
\toprule
\textbf{Name} & \textbf{Binder Length} & \textbf{Hemolysis}& \textbf{Non-Fouling} & \textbf{Solubility} & \textbf{Half-Life (h)} & \textbf{Affinity} \\
\midrule
AMHR2     & 8  & 0.9156 & 0.8613 & 0.8564 & 45.73& 7.0608 \\
AMHR2     & 12 & 0.9384 & 0.8872 & 0.8810 & 52.52& 7.2284 \\
AMHR2     & 16 & 0.9420 & 0.8914 & 0.8755 & 63.34& 7.2533 \\
EWS::FLI1  & 8  & 0.9186 & 0.8630 & 0.8619 & 44.77& 5.8424 \\
EWS::FLI1  & 12 & 0.9345& 0.8819& 0.8796& 59.11& 6.2007\\
EWS::FLI1  & 16 & 0.9416 & 0.8875 & 0.8807 & 64.32& 6.4195 \\
MYC       & 8  & 0.9180 & 0.8627 & 0.8627 & 44.13& 6.4082 \\
OX1R      & 10 & 0.9302 & 0.8687 & 0.8563 & 50.14& 7.1882 \\
DUSP12    & 9  & 0.9240 & 0.8669 & 0.8633 & 48.14& 6.1276 \\
1B8Q      & 8  & 0.9214 & 0.8680 & 0.8654 & 42.63& 5.7130 \\
5AZ8      & 11 & 0.9293 & 0.8732 & 0.8605 & 58.33& 6.2792 \\
7JVS      & 11 & 0.9313 & 0.8840 & 0.8743 & 56.49& 6.8449 \\
\bottomrule
\end{tabular}
\label{tab:5_classifiers}
\end{table}
\subsection{AReUReDi generates wild-type peptide binders under five property guidance}
We next benchmark AReUReDi on a wild-type peptide binder generation task guided by five different properties that are critical for therapeutic discovery: hemolysis, non-fouling, solubility, half-life, and binding affinity. To evaluate AReUReDi in a controlled setting, we designed 100 peptide binders per target for 8 diverse proteins, structured targets with known binders (3IDJ, 5AZ8, 7JVS), structured targets without known binders (AMHR2, OX1R, DUSP12), and intrinsically disordered targets (EWS::FLI1, MYC) (Table~\ref{tab:5_classifiers}). Across all targets and across multiple binder lengths, the generated peptides achieve superior hemolysis rates (0.91-0.94), high non‑fouling ($>$0.86) and solubility ($>$0.85), extended half‑life (42-64 h), and strong affinity scores (5.7-7.3), demonstrating both balanced optimization and robustness to sequence length. 

For the target proteins with pre-existing binders, we compared the property values between their known binders with AReUReDi-designed ones (Figure \ref{fig:visualization}A,B, \ref{fig:w_binders}). The designed binders significantly outperform the pre-existing binders across all properties without compromising the binding potential, which is further confirmed by the ipTM scores computed by AlphaFold3 \citep{abramson2024accurate} and docking scores calculated by AutoDock VINA \citep{trott2010autodock}. Although the AReUReDi-designed binders bind to similar target positions as the pre-existing ones, they differ significantly in sequence and structure, demonstrating AReUReDi’s capacity to explore the vast sequence space for optimal designs. For target proteins without known binders, complex structures were visualized using one of the AReUReDi-designed binders (Figure \ref{fig:w/o_binders}). The corresponding property scores, as well as ipTM and docking scores, are also displayed. Some of the designed binders showed longer half-life, while others excelled in non-fouling and solubility, underscoring the comprehensive exploration of the sequence space by AReUReDi.

To evaluate our guided generation strategy, we tracked the mean and standard deviation of five property scores across 100 generated binders (length 12) targeting EWS::FLI1 at each iteration (Figure \ref{fig:visualization}E). All five properties steadily improved, with average scores for solubility and non-fouling properties increasing markedly from \~0.4 to 0.9. The large standard deviation observed in the final half-life and binding affinity values reflects this property's high sensitivity to guidance, as AReUReDi balances the trade-offs between multiple conflicting objectives. We further visualized AReUReDi's impact by comparing the property distribution of the 100 guided peptides to that of 100 peptides unconditionally sampled from PepReDi$^3$ (Figure \ref{fig:visualization}F). The results show that AReUReDi effectively shifted the distribution towards peptides with higher binding affinity. Collectively, these findings demonstrate AReUReDi's capability to steer generation toward simultaneous multi-property optimization.

\renewcommand{\arraystretch}{1.3}
\begin{table}[t]
% \scriptsize
\midfootnotesize
\centering
\caption{AReUReDi outperforms traditional multi-objective optimization algorithms in designing wild-type peptide binders guided by five objectives. Each value represents the average of 100 designed binders. The table also records the average runtime for each algorithm to design a single binder. The best result for each metric is highlighted in bold.}
\label{tab:comparison}
% \vspace{0.5em}
\begin{tabular}{cccccccc}
\toprule
\textbf{Target} & \textbf{Method}   & \textbf{Time (s)} & \textbf{Hemolysis}& \textbf{Non-Fouling} & \textbf{Solubility} & \textbf{Half-Life (h)} & \textbf{Affinity} \\
\midrule
\multirow{6}{*}{1B8Q} 
  & MOPSO     & 8.54  & 0.8934 & 0.4763 & 0.4684 & 4.45& 6.0594 \\
  & NSGA-III  & 33.13 & 0.9138 & 0.5715 & 0.5825 & 7.32& 7.2178 \\
  & SMS-EMOA  & 8.21  & 0.8804 & 0.3450 & 0.3511 & 3.02& 5.955  \\
  & SPEA2     & 17.48 & 0.9181 & 0.4973 & 0.5057 & 4.13& \textbf{7.3240}\\
  & PepTune + DPLM & \textbf{2.46} & 0.8547 & 0.3085 & 0.3213 & 1.17& 5.2398 \\
  & \textbf{AReUReDi} & 55 & \textbf{0.9214} & \textbf{0.8680} & \textbf{0.8654} & \textbf{22.93}& 5.7130\\
\midrule
\multirow{6}{*}{PPP5} 
  & MOPSO     & 11.34 & 0.9117 & 0.4711 & 0.4255 & 1.77& 6.6958 \\
  & NSGA-III  & 37.30 & \textbf{0.9521} & 0.7138 & 0.7066 & 2.90& 7.3789 \\
  & SMS-EMOA  & 8.43  & 0.8758 & 0.4269 & 0.4334 & 1.03& 6.2854 \\
  & SPEA2     & 19.02 & 0.9445 & 0.6221 & 0.6098 & 2.61& \textbf{7.6253} \\
  & PepTune + DPLM & \textbf{4.80} & 0.8816 & 0.2752 & 0.2636 & 1.27& 5.8454 \\
  & \textbf{AReUReDi} & 195 & 0.9412 & \textbf{0.896} & \textbf{0.8832} & \textbf{38.28} & 6.7186 \\
\bottomrule
\end{tabular}
\end{table}

We benchmarked AReUReDi against four established multi-objective optimization (MOO) baselines (NSGA-III \citep{deb2013evolutionary}, SMS-EMOA \citep{beume2007sms}, SPEA2 \citep{zitzler2001spea2}, and MOPSO \citep{coello2002mopso}) on two protein targets: 1B8Q, a small protein with known peptide binders \citep{Zhang1999}, and PPP5, a larger protein without characterized binders \citep{Yang2004} (Table~\ref{tab:comparison}). Each method generated 100 candidate binders optimized for five properties: hemolysis, non-fouling, solubility, half-life, and binding affinity. While AReUReDi required longer runtimes than evolutionary baselines, it consistently produced the best trade-offs. For both targets, it designed targets with top hemolysis scores, increased non-fouling and solubility by 30-50\%, maintained competitive binding affinity, and even extended the half-life by a factor of 3-13 relative to the next-best method. These results underscore AReUReDi’s effectiveness in navigating high-dimensional property landscapes to yield peptide binders with balanced, optimized profiles.

We also compared against PepTune \citep{tang2025peptune}, a recent masked discrete diffusion model for peptide design that couples generation with Monte Carlo Tree Search for MOO. PepTune’s backbone was adapted to the existing DPLM model \citep{wang2024diffusion} for wild-type peptide sequence generation. Despite longer runtimes, AReUReDi substantially outperformed PepTune across all objectives, yielding nearly threefold improvements in non-fouling and solubility and a 22-fold increase in half-life. Together, these comparisons demonstrate that AReUReDi surpasses not only traditional MOO algorithms but also the current state-of-the-art diffusion-based approach for multi-objective-guided wild-type peptide binder design.

\begin{figure}[t]
    \centering
      \includegraphics[width=\textwidth]{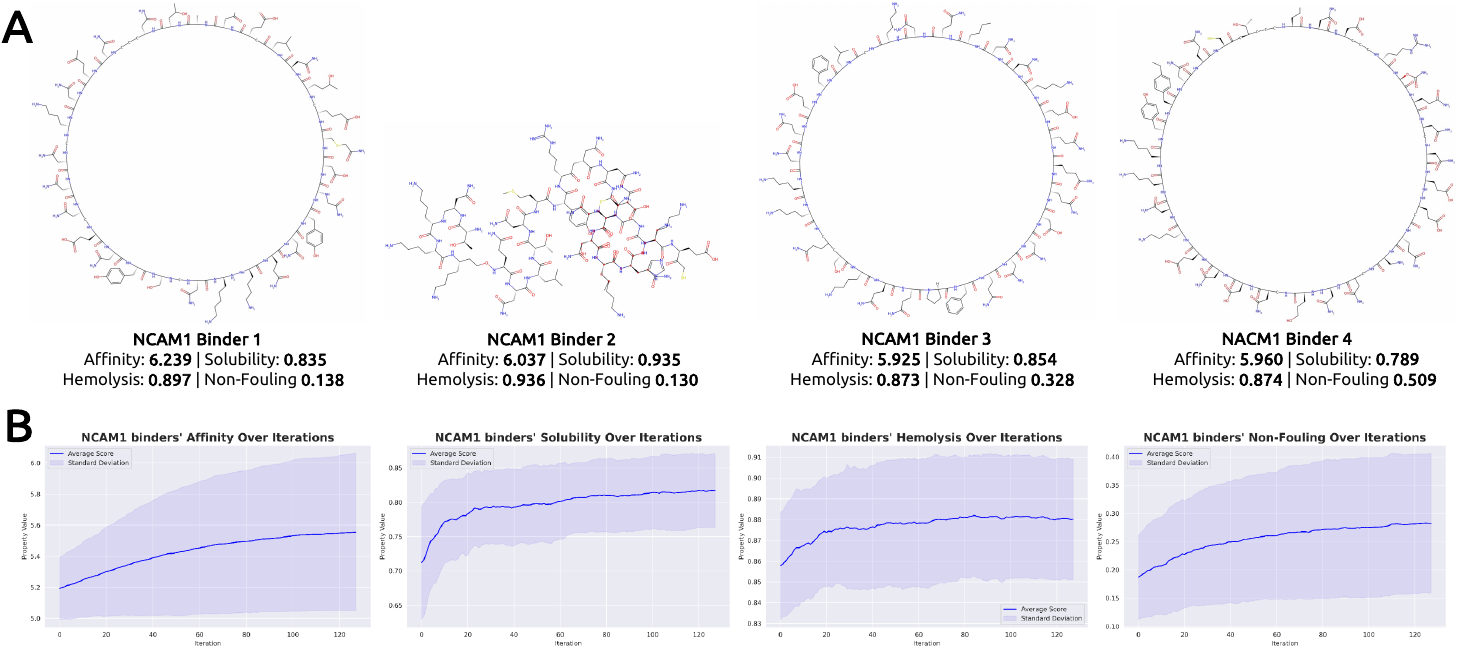}
    \caption{\textbf{(A)} Example 2D SMILES structure of AReUReDi-designed peptide binders with four property scores. \textbf{(B)} Plots showing the mean scores for each property across the number of iterations during AReUReDi's design of binders of length 200 for NCAM1.}
    \label{fig:NCAM1}
    \vskip -0.1in
\end{figure}

\subsection{AReUReDi generates therapeutic peptide SMILES under four property guidance}
To demonstrate the broad applicability of AReUReDi for multi-objective guided generation of biological sequences, we employed the rectified SMILESReDi model to design chemically-modified peptide binder SMILES sequences for five diverse therapeutic targets. These included the metabolic hormone receptor Glucagon-like peptide-1 receptor (GLP1), the iron transport protein Transferrin receptor (TfR), the Neural Cell Adhesion Molecule 1 (NCAM1), the neurotransmitter transporter GLAST, and the developmental Anti-Müllerian Hormone Receptor Type 2 (AMHR2). For each target, sequence generation was jointly conditioned on a predicted binding-affinity score to the target protein, as long as hemolysis, solubility, and non-fouling, to ensure both potency and desirable physicochemical profiles. Although PepTune is also able to perform multi-property guided design of peptide-binder SMILES sequences, it does not report average property scores for its generated binders, making a direct quantitative comparison with AReUReDi infeasible \citep{tang2025peptune}.

We selected and visualized representative binders with the highest predicted binding affinities for each target (Figure \ref{fig:NCAM1}A, \ref{fig:GLP1_GLAST}A,C, \ref{fig:TfR_AMHR2}A,C). All selected binders achieved high scores across hemolysis, solubility, non-fouling, and binding affinity. During generation, we recorded the mean and standard deviation of all four property scores over 100 binders at each iteration to assess the effectiveness of the multi-objective guidance (Figure \ref{fig:NCAM1}B, \ref{fig:GLP1_GLAST}B,D, \ref{fig:TfR_AMHR2}B,D). Across all targets, binding affinity scores and non-fouling scores showed steady upward trends throughout the generation process, while hemolysis and solubility scores fluctuated, indicating AReUReDi's effort to balance the four conflicting objectives. Moreover, AReUReDi produces valid sequences with substantially higher diversity and lower SNN than PepTune, indicating both superior novelty and structural variability (Table \ref{tab:SMILESReDi}). These findings highlight the versatility and reliability of AReUReDi for the \textit{de novo} design of chemically modified peptide binders across a wide range of therapeutic targets.

\subsection{Ablation Studies for Rectification and Annealed Guidance Strength }
To determine if rectification offers an advantage over standard discrete flow matching, we compared the performance of AReUReDi using three generative models: the base PepReDi model (no rectification), PepReDi (three rounds of rectification), and PepDFM, a standard discrete flow model that follows \citep{gat2024discrete} and was trained on the same data (Section \ref{sec:pepdfm}). Under the three settings, wild-type binders were designed for two distinct protein targets: 5AZ8 and AMHR2 (Table \ref{tab:ablation_rectification}). For the AMHR2 target, the rectified model achieved the highest scores across all five properties, with its predicted half-life surpassing the next-best method by nearly 13 hours. For the 5AZ8 target, the rectified model yielded a significantly higher half-life while maintaining comparable performance on other metrics. These results indicate that by lowering conditional TC and improving the quality of the probability path, rectification enables AReUReDi to achieve stronger Pareto trade-offs on the more demanding objectives.

We further demonstrated the advantage of using an annealed guidance strength (Table~\ref{tab:ablation_eta}). AReUReDi was applied to design wild-type peptide binders for two distinct proteins: a structured protein with known binders (PDB 1DDV) and an intrinsically disordered protein without known binders (P53). Across both targets, any fixed guidance strength, whether set to $\eta_{\min}$, $\eta_{\max}$, or their midpoint, failed to match the performance achieved with an annealed schedule. For 1DDV, annealing produced binders with markedly higher half-life and the best solubility, while maintaining hemolysis, non-fouling, and affinity scores that meet or exceed those of all fixed-$\eta$ settings. A similar trend holds for P53, where the annealing schedule consistently delivers the strongest results across all objectives. These findings confirm that gradually increasing the guidance strength enables AReUReDi to attain more favorable Pareto trade-offs, enhancing challenging properties such as half-life without sacrificing other therapeutic metrics.

\section{Related Works}

\textbf{Online Multi-Objective Optimization.}  
Recent work in multi-objective guided generation has focused on online or sequential decision-making, where solutions are refined with new data \citep{gruver2023protein, jain2023multi, stanton2022accelerating, ahmadianshalchi2024pareto}. A common approach is Bayesian optimization (BO), which builds a surrogate model and proposes evaluations via acquisition functions \citep{yu2020gradient, shahriari2015taking}. Multi-objective BO often uses advanced criteria such as EHVI \citep{emmerich2008computation}, information gain \citep{belakaria2021output}, or scalarization \citep{knowles2006parego, zhang2007moea, paria2020flexible}. While AReUReDi also employs Tchebycheff scalarization, it operates in an offline setting, where each sequence requires costly evaluation. This contrasts with the sequential, feedback-driven nature of online methods, making direct comparison inappropriate.

\textbf{Tchebycheff Scalarization.}  
Tchebycheff scalarization can identify any Pareto-optimal point and is widely used in multi-objective optimization \citep{Miettinen1999}. Recent variants include smooth scalarization for gradient-based algorithms \citep{lin2024smooth} and OMD-TCH for online learning \citep{liu2024online}. AReUReDi is, to our knowledge, the first to apply Tchebycheff scalarization for offline generative design of discrete therapeutic sequences. Future work may extend to many-objective problems or alternative utility functions \citep{lin2024few, tu2023multi}.

\textbf{Diffusion and Flow Matching.}  
Generative approaches such as ParetoFlow and PGD-MOO adapt flow matching or diffusion models for multi-objective optimization \citep{yuan2024paretoflow, annadani2025preference}. These operate in continuous or latent spaces, whereas AReUReDi is designed for discrete token spaces inherent to biological sequences. This domain mismatch precludes direct benchmarking.

\textbf{Biomolecule Generation.}  
Offline multi-objective frameworks such as EGD and MUDM have optimized molecules with multiple properties \citep{sun2025evolutionary, han2023training}, but these emphasize 3D structural representations. By contrast, AReUReDi is sequence-only, operating directly over amino acids or SMILES, which makes structural methods unsuitable as direct comparators.

\section{Discussion}
In this work, we have presented \textbf{AReUReDi}, a multi-objective optimization framework that extends rectified discrete flows to generate biomolecular sequences satisfying multiple, often conflicting, properties. By integrating annealed Tchebycheff scalarization, locally balanced proposals, and Metropolis-Hastings updates, AReUReDi provides theoretical guarantees of convergence to the Pareto front while maintaining full coverage of the solution space. Built on high-quality base generators such as PepReDi and SMILESReDi, the method demonstrates broad applicability across amino acid sequences and chemically modified peptide SMILES. Superior \textit{in silico} results establish AReUReDi as a general, theoretically-grounded tool for multi-property-guided biomolecular sequence design. 

While AReUReDi excels in domains like wild-type and chemically-modified peptide designs, future work will extend to other biological modalities, including DNA, RNA, antibodies, and combinatorial genotype libraries, where multi-objective trade-offs are central. From a theoretical perspective, improving AReUReDi's efficiency while maintaining the Pareto convergence guarantees and incorporating uncertainty-aware or feedback-driven guidance remain key directions to explore. Ultimately, AReUReDi provides a foundation for designing the next generation of therapeutic molecules that are not only potent but also explicitly optimized for the diverse properties required for clinical success.

\section*{Declarations}
\subsection*{Acknowledgments} We thank Mark III Systems for providing database and hardware support that has contributed to the research reported within this manuscript. We further thank Sophia Tang and Sophia Vincoff for assistance with PepTune benchmarking, and Qiang Liu for discussions related to Rectified Flow Matching. 

\subsection*{Author Contributions} T.C. designed and evaluated PepReDi, SMILESReDi, PepDFM, and AReUReDi, and performed model benchmarking and visualizations. Y.Z. curated and processed the PPIRef dataset for training, and performed molecular docking. P.C. conceived, designed, directed, and supervised the study.

\subsection*{Data and Materials Availability}
All sequences and data needed to evaluate the conclusions are presented in the paper and tables. All code to train PepReDi, SMILESReDi and generate sequences with AReUReDi are accessible by the academic community at \url{https://huggingface.co/ChatterjeeLab/AReUReDi}.

\subsection*{Competing Interests}
P.C. is a co-founder of Gameto, Inc., UbiquiTx, Inc., and Atom Bioworks, Inc., and advises companies involved in peptide development. P.C.’s interests are reviewed and managed by the University of Pennsylvania in accordance with their conflict-of-interest policies.

\clearpage
\bibliographystyle{apalike} 
\bibliography{AReUReDi}
\newpage

\newpage
\appendix
\section*{Supplementary Information}
% Reset figure counter and redefine figure numbering format for supplementary section
\setcounter{figure}{0} % Start figure numbering from 1
\setcounter{table}{0}
\setcounter{section}{0}
\renewcommand{\thefigure}{S\arabic{figure}} % Redefine figure numbering
\renewcommand{\thetable}{S\arabic{table}}
\renewcommand{\thesection}{S\arabic{section}}

\section{Theoretical Guarantees}\label{sec:proofs-main}
In this section, we establish that AReUReDi converges to Pareto-optimal
solutions while preserving coverage of the entire Pareto front. We assume throughout that the state space $\mathcal{S}$ is finite, all objective functions $s_n$ are bounded, and their normalized versions $\tilde{s}_n$ map to $[0,1]$.

\subsection{Preliminary Definitions}

\textbf{Definition (Pareto Optimality).}  
A state $x^* \in \mathcal{S}$ is \emph{Pareto optimal} if there exists no $y \in \mathcal{S}$ such that $\tilde{s}_n(y) \geq \tilde{s}_n(x^*)$ for all $n \in \{1,\ldots,N\}$ with strict inequality for at least one $n$.

\textbf{Definition (Pareto Front).}  
The Pareto front is $\mathcal{P} = \{x \in \mathcal{S} : x \text{ is Pareto optimal}\}$.

\textbf{Definition (Interior Weight Vector).}  
A weight vector $\omega \in \Delta^{N-1}$ is \emph{interior} if $\omega_n > 0$ for all $n$.

\subsection{Main Theoretical Results}

\textbf{Theorem (Invariance).}  
The Markov kernel defined by the Locally Balanced Proposal (LBP) and Metropolis--Hastings update leaves the distribution
\[
\pi_{\eta,\omega}(x) \propto p_1(x) \exp\!\big(\eta S_\omega(x)\big)
\]
invariant for every guidance strength $\eta>0$ and weight vector $\omega\in\Delta^{N-1}$.

\begin{proof}
We prove this in two steps: first showing that single-coordinate updates preserve detailed balance, then that random-scan mixtures preserve invariance.

\textbf{Step 1: Single-coordinate detailed balance.}  
Let $x$ and $x'$ differ only at coordinate $i$, where $x'_i = y$ for some token $y$. The proposal probability is
\[
q_i(y \mid x) = \frac{p_t^i(y \mid x_t) g(r_i(y; x_t))}{\sum_{z \in \text{candidates}} p_t^i(z \mid x_t) g(r_i(z; x_t))},
\]
where $r_i(y; x_t) = \frac{W_{\eta_t,\omega}(x_t^{(i \leftarrow y)})}{W_{\eta_t,\omega}(x_t)}$ and $g$ satisfies $g(u) = u \cdot g(1/u)$.

The acceptance probability is
\[
\alpha_i(x,x') = \min\left\{1, \frac{\pi_{\eta,\omega}(x') q_i(x_i \mid x')}{\pi_{\eta,\omega}(x) q_i(y \mid x)}\right\}.
\]

By the symmetry property of $g$ and the construction of the proposal, we have
\[
\frac{q_i(y \mid x)}{q_i(x_i \mid x')} = \frac{W_{\eta,\omega}(x')}{W_{\eta,\omega}(x)}.
\]

Since $\pi_{\eta,\omega}(x) = Z^{-1} p_1(x) W_{\eta,\omega}(x)$, it follows that
\[
\frac{\pi_{\eta,\omega}(x') q_i(x_i \mid x')}{\pi_{\eta,\omega}(x) q_i(y \mid x)} = 1.
\]

Therefore, $\alpha_i(x,x') = 1$ and detailed balance is satisfied.

\textbf{Step 2: Random-scan mixture.}  
The overall kernel is $K(x,x') = \frac{1}{L} \sum_{i=1}^L K_i(x,x')$, where $K_i$ is the kernel for updating coordinate $i$. Since each $K_i$ satisfies detailed balance with respect to $\pi_{\eta,\omega}$, their convex combination also satisfies detailed balance and hence preserves invariance.
\end{proof}

\textbf{Theorem (Convergence to Pareto Front).}  
Fix any $\omega \in \mathrm{int}\,\Delta^{N-1}$ with strictly positive entries and let $S_\omega(x) = \min_n \omega_n \tilde{s}_n(x)$. If $\eta \to \infty$, samples drawn from $\pi_{\eta,\omega}(x) \propto p_1(x) \exp(\eta S_\omega(x))$ concentrate on the set
\[
\mathcal{F}_\omega = \arg\max_x S_\omega(x),
\]
and every element of $\mathcal{F}_\omega$ is Pareto optimal.

\begin{proof}
\textbf{Step 1: Maximizers of $S_\omega$ are Pareto optimal.}  
Suppose $x^* \in \mathcal{F}_\omega$ but $x^*$ is not Pareto optimal. Then there exists $y \in \mathcal{S}$ with
\[
\tilde{s}_n(y) \geq \tilde{s}_n(x^*) \ \forall n, \quad \text{and} \quad \tilde{s}_m(y) > \tilde{s}_m(x^*) \ \text{for some } m.
\]
Since $\omega_n > 0$ for all $n$, multiplying preserves inequalities. If $m$ is the bottleneck coordinate of $x^*$, then $S_\omega(y) > S_\omega(x^*)$, contradiction. Otherwise, equality requires special weight alignments (measure zero). Thus maximizers are Pareto optimal almost surely.

\textbf{Step 2: Concentration as $\eta \to \infty$.}  
Let $S_\omega^* = \max_x S_\omega(x)$ and $\Delta_\omega = S_\omega^* - \max_{x \notin \mathcal{F}_\omega} S_\omega(x) > 0$. Then for $x \notin \mathcal{F}_\omega$,
\[
\pi_{\eta,\omega}(x) \leq e^{-\eta \Delta_\omega}\cdot\frac{p_1(x)}{\sum_{z \in \mathcal{F}_\omega} p_1(z)}.
\]
Summing gives $\pi_{\eta,\omega}(\mathcal{S}\setminus\mathcal{F}_\omega) \to 0$ as $\eta \to \infty$. Hence the mass concentrates on $\mathcal{F}_\omega$.
\end{proof}

\textbf{Theorem (Pareto Point Representability).}  
For every Pareto-optimal state $x^\dagger \in \mathcal{P}$ there exists $\omega \in \Delta^{N-1}$ such that $x^\dagger \in \arg\max_x S_\omega(x)$. Moreover, if $\tilde{s}_n(x^\dagger) > 0$ for all $n$, then $x^\dagger$ can be made the unique maximizer.

\begin{proof}
If $\tilde{s}_n(x^\dagger) > 0$, define
\[
\omega_n = \frac{1/\tilde{s}_n(x^\dagger)}{\sum_{k=1}^N 1/\tilde{s}_k(x^\dagger)}.
\]
Then $S_\omega(x^\dagger) = \frac{1}{\sum_k 1/\tilde{s}_k(x^\dagger)}$, and for any $y \neq x^\dagger$, some $m$ satisfies $\tilde{s}_m(y) < \tilde{s}_m(x^\dagger)$, implying $S_\omega(y) < S_\omega(x^\dagger)$.  
If some $\tilde{s}_n(x^\dagger) = 0$, perturb objectives by $\varepsilon > 0$ and take the limit.
\end{proof}

\textbf{Theorem (Coverage Guarantee).}  
Let $\mu$ be any probability distribution with full support on $\mathrm{int}\,\Delta^{N-1}$. If $\omega \sim \mu$ and $\eta \to \infty$, then the induced sampler visits every Pareto-optimal state with positive probability.

\begin{proof}
By representability, each Pareto point $x^\dagger$ maximizes $S_\omega$ for some interior $\omega^\dagger$. By continuity, there exists a neighborhood $U_{x^\dagger}$ where $x^\dagger$ remains optimal. Since $\mu(U_{x^\dagger}) > 0$, randomizing $\omega$ ensures $x^\dagger$ is visited with positive probability in the high-$\eta$ limit.
\end{proof}

\textbf{Remark.}  
The guarantees hold for any finite $\mathcal{S}$ and bounded objectives. In practice, convergence depends on the chain mixing rate, the annealing schedule for $\eta$, and the choice of balancing function $g$.

\section{Base Model Details}
\subsection{PepReDi}
\textbf{Model Architecture.} The backbone of PepReDi is built on a Diffusion Transformer (DiT) framework implemented within a Masked Diffusion Language Model (MDLM) paradigm \citep{Peebles2022DiT, sahoo2024mdlm}. Input amino acid sequences are transformed to discrete tokens using the ESM-2-650M tokenizer \citep{lin2023evolutionary}. Tokenized amino acid sequences and time-steps are converted to continuous embedding vectors using two separate layers, which are then fused and processed by stacked DiT transformer blocks equipped with multi-head self-attention to capture long-range dependencies in the amino-acid sequence. Residual connections and layer normalization stabilize the training dynamics, and a final projection layer outputs token logits for each position.

\textbf{Dataset Curation.} The dataset for PepReDi training was curated from the PepNN, BioLip2, and PPIRef dataset \citep{abdin2022pepnn, zhang2024biolip2, bushuiev2023learning}. All peptides from PepNN and BioLip2 were included, along with sequences from PPIRef ranging from 6 to 49 amino acids in length. The dataset was divided into training, validation, and test sets at an 80/10/10 ratio.

\textbf{Training Strategy. }Training was conducted on a single node equipped with one NVIDIA GPU and 128 GB of GPU memory using the SLURM workload manager. The model was trained for 100 epochs using the Adam optimizer and a learning rate of 1e-4 with weight decay of 1e-5. A learning rate scheduler with 10 warm-up epochs and cosine decay was used, with initial and minimum learning rates both 1e-5. The network architecture included a model dimension of 512, 6 transformer layers, and 8 attention heads, with a vocabulary size of 24 and a maximum sequence length of 100 tokens. Conditional total correlation estimation was performed using 20 batches and 50 samples per batch to monitor rectification quality during training. The model checkpoint with the lowest total correlation was saved. For training rectified models, the same hyperparameter setting was applied, except for the loaded pre-trained model checkpoint and the weight decay being increased to 2e-5.

\textbf{Dynamic Batching.} To enhance computational efficiency and manage variable-length token sequences, we implemented dynamic batching. Drawing inspiration from ESM-2’s approach \citep{lin2023evolutionary}, input peptide sequences were sorted by length to optimize GPU memory utilization, with a maximum token size of 100 per GPU.

\textbf{Rectification. } The trained model applied 16 sampling steps to generate 10k sequences for each peptide length, ranging from 6 to 49, with a temperature hyperparameter set to 1. After generation, dynamic batching was used to optimize GPU memory utilization for future rectified training.

\subsection{SMILESReDi}
\label{sec:SMILESReDi}
\textbf{Model Architecture. }SMILESReDi follows the ReDi paradigm and uses a Diffusion Transformer (DiT) backbone embedded in a Masked Diffusion Language Model (MDLM) design to generate molecular SMILES sequences \citep{Peebles2022DiT, sahoo2024mdlm}. Input SMILES sequences are transformed to discrete tokens using the PeptideCLM -23M tokenizer. Tokenized amino acid sequences and time-steps are converted to continuous embedding vectors using two separate layers. Both embeddings are then fused and processed by stacked DiT transformer blocks that incorporate Rotary Positional Embeddings (RoPE) and multi-head attention modules to capture long-range structural dependencies while preserving positional information \citep{su2024roformer}. A final layer normalization and linear projection outputs token logits for each position.

\textbf{Time-dependent bond-aware noising schedule. }Peptide SMILES share a conserved backbone of alternating carbonyl and amide groups connected by chemically constrained peptide bonds, while their side chains remain highly diverse. Standard discrete flow matching can corrupt these critical bond tokens too early, hindering the flow from recovering the backbone along the probability path. Inspired by previous work in bond-dependent masking, we devised a time-dependent bond-aware noising schedule that preserves backbone tokens longer than side-chain tokens, allowing the model to reconstruct the invariant scaffold before generating variable side chains. Specifically, for each position $j$ with a bond indicator $b_j\in\{0,1\}$, the time-$t$ marginal of the probability path is
\[
p_t(x_t^{(j)} \mid x_0^{(j)},x_1^{(j)})
   = \big[\,b_j t^{\gamma} + (1-b_j)t\,\big]\,
      \delta_{x_1^{(j)}} +
     \big[\,1 - b_j t^{\gamma} - (1-b_j)t\,\big]\,
      \delta_{x_0^{(j)}},
\quad t\in[0,1],\ \gamma>1,
\]
so each token is equal to $x_1^{(j)}$ with the indicated mixture coefficient and to $x_0^{(j)}$ otherwise, ensuring that backbone tokens ($b_j=1$) transition more slowly than non-bond tokens along the DFM probability path.

\textbf{Training Strategy. } The training is conducted on a 4*A6000 NVIDIA RTX 6000 Ada GPU system with 48 GB of VRAM for 5 epochs. The model checkpoint with the lowest evaluation loss was saved. The Adam optimizer was employed with a learning rate of 1e-4. A learning rate scheduler with 10\% total training steps and cosine decay was used, with initial and minimum learning rates both 1e-5. The network architecture included a model dimension of 768, 8 transformer layers, and 8 attention heads. Gradient clip value was set to 1.0 and $\gamma$ to 2.0 in the time-dependent bond-aware noising schedule. For training rectified models, the same hyperparameter setting was applied, except for the loaded pre-trained model checkpoint and the total training epochs set to 10. 

\textbf{Rectification. } The trained model applied 100 sampling steps to generate 100 sequences for each peptide length, ranging from 4 to 1035, with a temperature hyperparameter set to 1. After generation, dynamic batching was used to optimize GPU memory utilization for future rectified training.

\textbf{Evaluation Metrics. }
\begin{itemize}
    \item \textbf{Validity} is defined as the fraction of peptide SMILES that pass the SMILES2PEPTIDE filter \citep{tang2025peptune}, indicating that it translates to a synthesizable peptide.
    \item \textbf{Uniqueness} is defined as the fraction of mutually distinct peptide SMILES.
    \item \textbf{Diversity} is defined as one minus the average Tanimoto similarity between the Morgan fingerprints of every pair of generated sequences, which measures the similarity in structure across generated peptides.
$$
\text{Diversity} = 1 - \frac{1}{\binom{N_{\text{generated}}}{2}} \sum_{i,j} \frac{\mathbf{f}(\mathbf{x}_i) \cdot \mathbf{f}(\mathbf{x}_j)}{|\mathbf{f}(\mathbf{x}_i)| + |\mathbf{f}(\mathbf{x}_j)| - \mathbf{f}(\mathbf{x}_i) \cdot \mathbf{f}(\mathbf{x}_j)}
$$
where $\mathbf{f}(\mathbf{x}_i)$ and $\mathbf{f}(\mathbf{x}_j)$ are the 2048-dimensional Morgan fingerprint with radius 3 for a pair of generated sequences $\mathbf{x}_i$ and $\mathbf{x}_j$.
\item \textbf{Similarity to Nearest Neighbor (SNN)} is defined as the maximum Tanimoto similarity between a generated sequence $\mathbf{x}_i$ with a sequence in the dataset $\tilde{\mathbf{x}}_j$.
$$
\text{SNN} = \max_{j \in |\mathcal{D}|} \left( \frac{\mathbf{f}(\mathbf{x}_i) \cdot \mathbf{f}(\tilde{\mathbf{x}}_j)}{|\mathbf{f}(\mathbf{x}_i)| + |\mathbf{f}(\tilde{\mathbf{x}}_j)| - \mathbf{f}(\mathbf{x}_i) \cdot \mathbf{f}(\tilde{\mathbf{x}}_j)} \right)
$$
\end{itemize}

\subsection{PepDFM}
\label{sec:pepdfm}
\textbf{Model Architecture.} The base model is a time-dependent architecture based on U-Net \citep{ronneberger2015u}. It uses two separate embedding layers for sequence and time, followed by five convolutional blocks with varying dilation rates to capture temporal dependencies, while incorporating time-conditioning through dense layers.  The final output layer generates logits for each token. We used a polynomial convex schedule with a polynomial exponent of 2.0 for the mixture discrete probability path in the discrete flow matching. 

\textbf{Dataset Curation.} The dataset for PepDFM training was curated from the PepNN, BioLip2, and PPIRef dataset \citep{abdin2022pepnn, zhang2024biolip2, bushuiev2023learning}. All peptides from PepNN and BioLip2 were included, along with sequences from PPIRef ranging from 6 to 49 amino acids in length. The dataset was divided into training, validation, and test sets at an 80/10/10 ratio.

\textbf{Training Strategy.} The training is conducted on a 2xH100 NVIDIA NVL GPU system with 94 GB of VRAM for 200 epochs with batch size 512. The model checkpoint with the lowest evaluation loss was saved. The Adam optimizer was employed with a learning rate 1e-4. A learning rate scheduler with 20 warm-up epochs and cosine decay was used, with initial and minimum learning rates both 1e-5. The embedding dimension and hidden dimension were set to be 512 and 256 respectively for the base model.

\textbf{Performance. }PepDFM achieved a validation loss of 3.1051. Its low generalized KL loss during evaluation demonstrates PepDFM's strong capability to generate sequences with high biological plausibility \citep{gat2024discrete}. 

\section{Objective Description}
In this work, five key property objectives are considered in the peptide binder tasks: hemolysis, non-fouling, solubility, half-life, and binding affinity. Each of these properties plays a crucial role in optimizing the therapeutic potential of peptides. Hemolysis refers to the peptide’s ability to minimize red blood cell lysis, ensuring safe systemic circulation \citep{pirtskhalava2013transmembrane}. Non-fouling properties describe the peptide’s resistance to unwanted interactions with biomolecules, thus enhancing its stability and bioavailability in vivo \citep{chen2009ultra}. Solubility is critical for ensuring adequate peptide dissolution in biological fluids, directly influencing its absorption and therapeutic efficacy \citep{fosgerau2015peptide}. Half-life indicates the duration for which the peptide remains active in circulation, which is vital for reducing dosing frequency \citep{swanson2014long}. Finally, binding affinity measures the strength of the peptide’s interaction with its target, directly correlating to its biological activity and potency in therapeutic applications \citep{bostrom2008improving}.

\section{Score Model Details}
\label{sec:score_models}
We applied the score models from \citep{tang2025peptune} to guide the generation of chemically-modified peptide binders. We now introduce the score model developed for the wild-type peptide binder generation task.  We collected hemolysis (9,316), non-fouling (17,185), solubility (18,453), and binding affinity (1,781) data for classifier training from the PepLand and PeptideBERT datasets \citep{zhang2023pepland, guntuboina2023peptidebert}. All sequences taken are wild-type L-amino acids and are tokenized and represented by the ESM-2 protein language model \citep{lin2023evolutionary}. 

\subsection{Boosted Trees for Classification}
For hemolysis, non-fouling, and solubility classification, we trained XGBoost boosted tree models for logistic regression. We split the data into 0.8/0.2 train/validation using stratified splits from scikit-learn \citep{scikit-learn} and generated mean-pooled ESM-2-650M \citep{lin2023evolutionary} embeddings as input features to the model. We ran 50 trials of OPTUNA \citep{akiba2019optuna} search to determine the optimal XGBoost hyperparameters (Table \ref{tab:xgboost_classification_params}), tracking the best binary classification F1 scores. The best models for each property reached F1 scores of 0.58, 0.71, and 0.68 on the validation sets respectively.

\begin{table}[ht]
\centering
\caption{XGBoost Hyperparameters for Classification}
\label{tab:xgboost_classification_params}
\vskip 0.05in
\begin{tabular}{@{}ll@{}}
\toprule
\textbf{Hyperparameter}       & \textbf{Value/Range}          \\ \midrule
Objective                     & \texttt{binary:logistic}      \\
Lambda                        & \([1\text{e}{-8}, 10.0]\)     \\
Alpha                         & \([1\text{e}{-8}, 10.0]\)     \\
Colsample by Tree             & \([0.1, 1.0]\)               \\
Subsample                     & \([0.1, 1.0]\)               \\
Learning Rate                 & \([0.01, 0.3]\)              \\
Max Depth                     & \([2, 30]\)                  \\
Min Child Weight              & \([1, 20]\)                  \\
Tree Method                   & \texttt{hist}                \\
\bottomrule
\end{tabular}
\end{table}

\subsection{Binding Affinity Score Model}
We developed an unpooled reciprocal attention transformer model to predict protein-peptide binding affinity, leveraging latent representations from the ESM-2 650M protein language model \citep{lin2023evolutionary}. Instead of relying on pooled representations, the model retains unpooled token-level embeddings from ESM-2, which are passed through convolutional layers followed by cross-attention layers. The binding affinity data were split into a 0.8/0.2 ratio, maintaining similar affinity score distributions across splits. We used OPTUNA \citep{akiba2019optuna} for hyperparameter optimization, tracing validation correlation scores. The final model was trained for 50 epochs with a learning rate of 3.84e-5, a dropout rate of 0.15, 3 initial CNN kernel layers (dimension 384), 4 cross-attention layers (dimension 2048), and a shared prediction head (dimension 1024) in the end. The classifier reached 0.64 Spearman's correlation score on validation data.

\subsection{Half-Life Score Model}
\textbf{Dataset Curation.} The half-life dataset is curated from three publicly available datasets: PEPLife, PepTherDia, and THPdb2 \citep{mathur2016peplife,d2021peptherdia,jain2024thpdb2}. Data related to human subjects were selected, and entries with missing half-life values were excluded. After removing duplicates, the final dataset consists of 105 entries.

\textbf{Pre-training on stability data.} Given the small size of the half-life dataset, which is insufficient for training a model to capture the underlying data distribution, we first pre-trained a score model on a larger stability dataset to predict peptide stability \citep{tsuboyama2023mega}. The model consists of three linear layers with ReLU activation functions, and a dropout rate of 0.3 was applied. The model was trained on a 2xH100 NVIDIA NVL GPU system with 94 GB of VRAM for 50 epochs. The Adam optimizer was employed with a learning rate of 1e-2. A learning rate scheduler with 5 warm-up epochs and cosine decay was used, with initial and minimum learning rates both 1e-3. After training, the model achieved a validation Spearman's correlation of 0.7915 and an $R^2$ value of 0.6864, demonstrating the reliability of the stability score model.

\textbf{Fine-tuning on half-life data.} The pre-trained stability score model was subsequently fine-tuned on the half-life dataset. Since half-life values span a wide range, the model was adapted to predict the base-10 logarithm of the half-life (h) values to stabilize the learning process. After fine-tuning, the model achieved a validation Spearman's correlation of 0.8581 and an $R^2$ value of 0.5977.

\section{Sampling Details}
\textbf{Score Model Settings.} We cap the predicted log-scale half-life at 2 (i.e., 100 h) to prevent it from dominating the optimization and ensure balanced trade-offs across all properties. For the remaining objectives, hemolysis, non-fouling, solubility, and binding affinity, we directly employ their model outputs during sampling.

\textbf{Wild-Type Peptide Binder Generation Task Settings.} The total sampling steps are set to 20 multiplied by the binder length. All possible candidate token transitions are evaluated during each sampling step. We applied the same weight for each objective in all wild-type peptide binder generation tasks.

\textbf{Chemically-Modified Peptide Binder Generation Task Settings.} The total sampling steps are set to 128. With a vocabulary size of 586, evaluating all the possible candidate tokens is too computationally intensive. We therefore only evaluated the top 200 candidate tokens during each sampling step. We applied weight 0.7 for binding affinity, and 0.1 for hemolysis, non-fouling, and solubility, respectively. Instead of random initialization, the initial sequences $x_0$ are sampled from the pre-trained SMILESReDi$^1$ with 16 generation steps. During generation, AReUReDi rejects any transitions that will make the SMILES sequence an invalid peptide.

\renewcommand{\arraystretch}{1.3}
\begin{table}[t]
% \scriptsize
\midfootnotesize
\centering
\caption{\textbf{Adding a sampling constraint greatly improves AReUReDi's performance.} Wild-type binders for two protein targets (PDB 8CN1 and 4EBP2) were generated with or without a sampling constraint using the same number of generation steps. The table reports the average score for each objective, calculated from 100 generated binders per setting. The best score for each objective is highlighted in bold.}
\label{tab:ablation_constraint}
% \vspace{0.5em}
\begin{tabular}{ccccccc}
\toprule
\textbf{Target} & \textbf{Method}   & \textbf{Hemolysis}& \textbf{Non-Fouling} & \textbf{Solubility} & \textbf{Half-Life} & \textbf{Affinity} \\
\midrule
\multirow{2}{*}{8CN1}& w/o constraints& 0.8650& 0.4782& 0.4627& 2.54& 5.2412\\
  & w/  constraints& \textbf{0.9213}& \textbf{0.8676}& \textbf{0.8697}& \textbf{44.70}& \textbf{5.5143}\\
\midrule
\multirow{2}{*}{4EBP2}& w/o constraints
& 0.8879& 0.4288& 0.4257& 1.8781& 5.7132\\
  & w/  constraints& \textbf{0.9356}& \textbf{0.8767}& \textbf{0.8692}& \textbf{53.95}& \textbf{6.4571}\\
\bottomrule
\end{tabular}
\end{table}

\renewcommand{\arraystretch}{1.5}
\begin{table}[ht]
\centering
\caption{Ablation results for wild-type peptide binder design targeting PDB 7LUL with different guidance settings. For each setting, 100 binders of length 7 were designed.}
\vspace{0.5em}
\begin{tabular}{l|ccc}
\hline
\textbf{\makecell[c]{Guidance Settings\\[0.4ex]Hemolysis\hspace{0.5em}Solubility\hspace{0.5em}Affinity}} & \textbf{Hemolysis} \rule{0pt}{1.6em}& \textbf{Solubility} & \textbf{Affinity}\\ 
\hline
\makecell[l]{\hspace{2em}$\checkmark$ \hspace{3em} $\checkmark$ \hspace{3em} $\checkmark$}& 0.9389& 0.9398& 6.2559\\
\hline
\makecell[l]{\hspace{2em}$\times$ \hspace{3em} $\checkmark$ \hspace{3em} $\checkmark$} & 0.8964& 0.9465& 6.3272\\
\hline
\makecell[l]{\hspace{2em}$\checkmark$ \hspace{3em} $\times$ \hspace{3em} $\checkmark$} & 0.9502& 0.4013& 6.9798\\
\hline
\makecell[l]{\hspace{2em}$\checkmark$ \hspace{3em} $\checkmark$ \hspace{3em} $\times$} & 0.9535& 0.9642& 5.2611\\
\hline
\makecell[l]{\hspace{2em}$\times$ \hspace{3em} $\times$ \hspace{3em} $\checkmark$} & 0.8812& 0.2877& 7.5057\\
\hline
\makecell[l]{\hspace{2em}$\times$ \hspace{3em} $\checkmark$ \hspace{3em} $\times$} & 0.9036& 0.9725& 5.2449\\
\hline
\makecell[l]{\hspace{2em}$\checkmark$ \hspace{3em} $\times$ \hspace{3em} $\times$} & 0.9802& 0.6135& 5.0985\\
\hline
\makecell[l]{\hspace{2em}$\times$ \hspace{3em} $\times$ \hspace{3em} $\times$} & 0.8431& 0.5810& 4.8919\\
\hline
\end{tabular}

\label{tab:ablation_7LUL}
\end{table}
\renewcommand{\arraystretch}{1.5}
\begin{table}[ht]
\centering
\caption{Ablation results for wild-type peptide binder design targeting PDB CLK1 with different guidance settings. For each setting, 100 binders of length 12 were designed.}
\vspace{0.5em}
\begin{tabular}{l|ccc}
\hline
\textbf{\makecell[c]{Guidance Settings\\[0.4ex]Non-Fouling\hspace{0.5em}Half-Life (h) \hspace{0.5em}Affinity}} 
  & \textbf{Non‑Fouling} \rule{0pt}{1.6em}& \textbf{Half‑Life (h)}& \textbf{Affinity}\\
\hline
\makecell[l]{\hspace{2em} $\checkmark$ \hspace{4em}
             $\checkmark$ \hspace{4.2em}
            $\checkmark$}& 0.8285& 74.04& 6.8099\\
\hline
\makecell[l]{\hspace{2em} $\times$ \hspace{4em}
            $\checkmark$ \hspace{4.2em}
            $\checkmark$}& 0.2902& 96.59& 7.3906\\
\hline
\makecell[l]{\hspace{2em} $\checkmark$ \hspace{4em}
             $\times$ \hspace{4.2em}
             $\checkmark$}& 0.9365& 1.33& 7.2029\\
\hline
\makecell[l]{\hspace{2em} $\checkmark$ \hspace{4em}
             $\checkmark$ \hspace{4.2em} 
             $\times$}
  & 0.9479& 75.68&  6.3437\\
\hline
\makecell[l]{\hspace{2em} $\times$ \hspace{4em}
             $\times$ \hspace{4.2em} 
             $\checkmark$}
  & 0.9625& 1.23& 6.2319\\
\hline
\makecell[l]{\hspace{2em} $\times$ \hspace{4em}
             $\checkmark$ \hspace{4.2em} 
             $\times$}
  & 0.3540& 100.00&  6.4116\\
\hline
\makecell[l]{\hspace{2em} $\checkmark$ \hspace{4em}
             $\times$ \hspace{4.2em} 
             $\times$}
  & 0.2531& 2.96&  8.6580\\
\hline
\makecell[l]{\hspace{2em} $\times$ \hspace{4em}
             $\times$ \hspace{4.2em} 
             $\times$}
  & 0.4988& 1.82&  5.4739\\ 
\hline

\end{tabular}
\label{tab:ablation_CLK1}
\end{table}

\renewcommand{\arraystretch}{1.3}
\begin{table}[t]
% \scriptsize
\midfootnotesize
\centering
\caption{\textbf{Rectification of the base generation model improves AReUReDi's performance.} Wild-type binders for two protein targets (PDB 5AZ8 and AMHR2)  were generated using AReUReDi with three different base models: PepDFM, PepReDi (without rectification), and PepReDi$^3$ (with three rounds of rectification). The table reports the average score for each objective, calculated from 100 generated binders per setting. The best score for each objective is highlighted in bold.}
\label{tab:ablation_rectification}
% \vspace{0.5em}
\begin{tabular}{ccccccc}
\toprule
\textbf{Target} & \textbf{Base Model}& \textbf{Hemolysis}& \textbf{Non-Fouling} & \textbf{Solubility} & \textbf{Half-Life} & \textbf{Affinity} \\
\midrule
\multirow{3}{*}{5AZ8}& PepDFM& 0.9296& \textbf{0.8867}& \textbf{0.8743}& 37.30& 6.2291\\
 & PepReDi& \textbf{0.9326}& 0.8759& 0.8572& 50.16&\textbf{6.4391}\\
  & PepReDi$^3$& 0.9293& 0.8732& 0.8605& \textbf{58.33}& 6.2792\\
\midrule
\multirow{3}{*}{AMHR2}& PepDFM
& 0.9412& 0.8774& 0.8612& 47.84& 7.2373\\
 & PepReDi
& 0.9127& 0.8602& 0.8460& 50.92&7.0101\\
  & PepReDi$^3$& \textbf{0.9420}& \textbf{0.8914}& \textbf{0.8755}& \textbf{63.34}& \textbf{7.2533}\\
\bottomrule
\end{tabular}
\end{table}

\renewcommand{\arraystretch}{1.3}
\begin{table}[t]
% \scriptsize
\midfootnotesize
\centering
\caption{\textbf{Annealed guidance strength improves AReUReDi's performance.} Wild-type binders for two protein targets (PDB 1DDV and P53) were generated under four guidance schedules: (1) fixed at the minimum strength $\eta_{min}=1.0$, (2) fixed at the maximum strength $\eta_{max}=20.0$, (3) fixed at the midpoint $\frac{1}{2}(\eta_{min} +\eta_{max})=10.5$, and (4) an annealed schedule where $\eta_t$ increases from $\eta_{min}$ to $\eta_{max}$ over optimization steps. The table reports the average score for each objective, calculated from 100 generated binders per setting. The best score for each objective is highlighted in bold.}
\label{tab:ablation_eta}
% \vspace{0.5em}
\begin{tabular}{ccccccc}
\toprule
\textbf{Target} & \textbf{Method}   & \textbf{Hemolysis}& \textbf{Non-Fouling} & \textbf{Solubility} & \textbf{Half-Life (h)} & \textbf{Affinity} \\
\midrule
\multirow{4}{*}{1DDV}& $\eta = \eta_{min}$& 0.9130& 0.8575& 0.8429& 38.70& 5.3554\\
  & $\eta = \eta_{max}$& 0.9156& 0.8512& 0.8479& 40.27& 5.4359\\
 & $\eta = \frac{1}{2}(\eta_{min} +\eta_{max})$& 0.9108& 0.8641& 0.8544& 40.43&5.5396\\
 & $\eta_t \;=\; \eta_{\min}+\bigl(\eta_{\max}-\eta_{\min}\bigr)\frac{t}{\,T-1\,}$ & 0.9128& 0.8545& \textbf{0.8565}& \textbf{44.73}&5.4482\\
\midrule
\multirow{4}{*}{P53}& $\eta = \eta_{min}$& 0.9335& 0.8800& 0.8706& 49.97& 6.2538\\
 & $\eta = \eta_{max}$& 0.9293& 0.8693& 0.8657& 61.76&6.3043\\
 & $\eta = \frac{1}{2}(\eta_{min} +\eta_{max})$& 0.9294& 0.8713& 0.8653& 59.43&6.3060\\
  & $\eta_t \;=\; \eta_{\min}+\bigl(\eta_{\max}-\eta_{\min}\bigr)\frac{t}{\,T-1\,}$ & \textbf{0.9353}& \textbf{0.8818}& \textbf{0.8785}& \textbf{62.83}& \textbf{6.3508}\\
\bottomrule
\end{tabular}
\end{table}

\begin{figure}
    \centering
    \includegraphics[width=\textwidth]{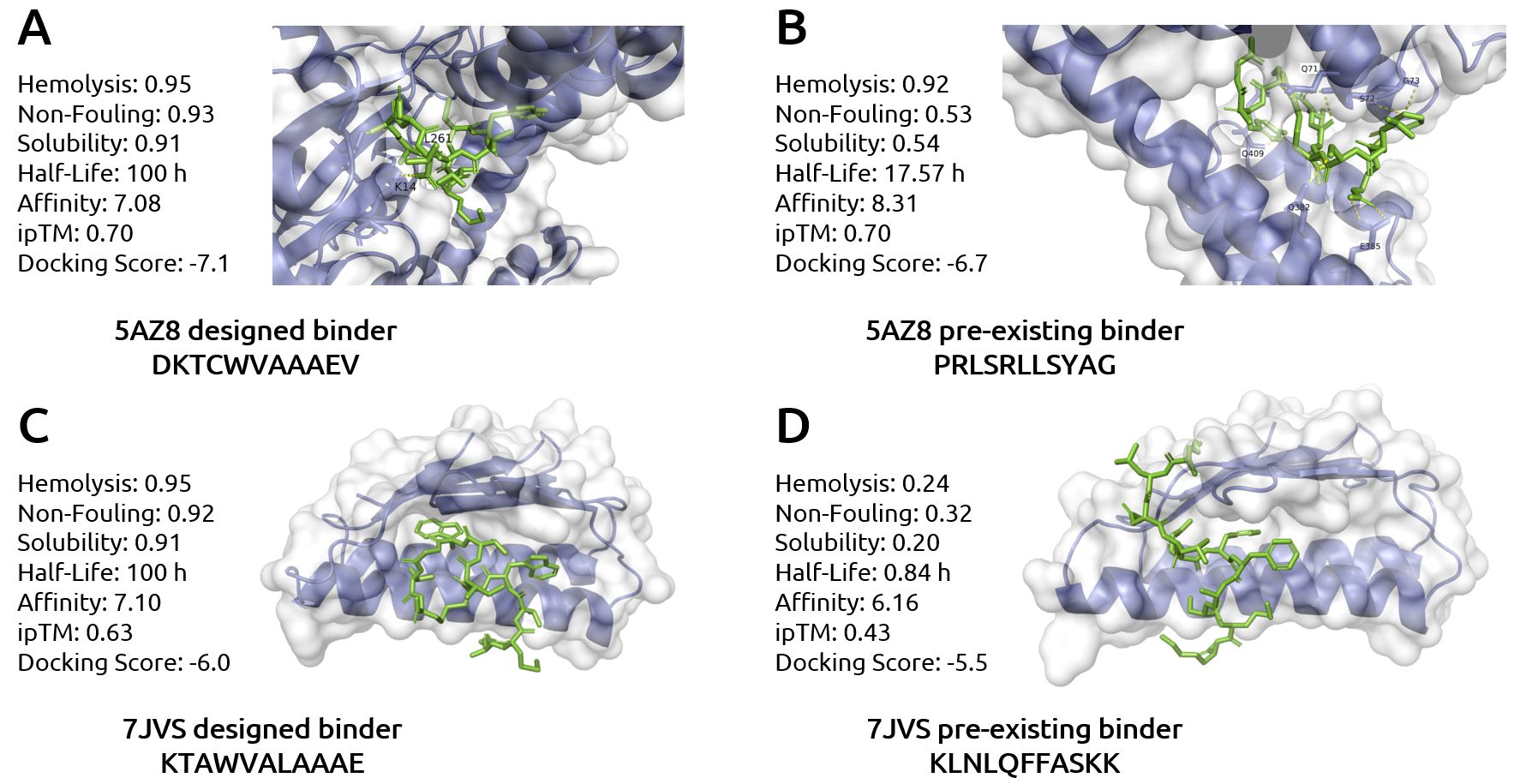}
    \caption{\textbf{Complex structures of target proteins with pre-existing binders.} \textbf{(A)-(B)} 5AZ8 \textbf{(C)-(D)} 7JVS. Each panel shows the complex structure of the target with either an AReUReDi-designed binder or its pre-existing binder. For each binder, five property scores are provided, as well as the ipTM score from AlphaFold3 and the docking score from AutoDock VINA. Interacting residues on the target are visualized.}
    \label{fig:w_binders}
\end{figure}

\begin{figure}
    \centering
    \includegraphics[width=\textwidth]{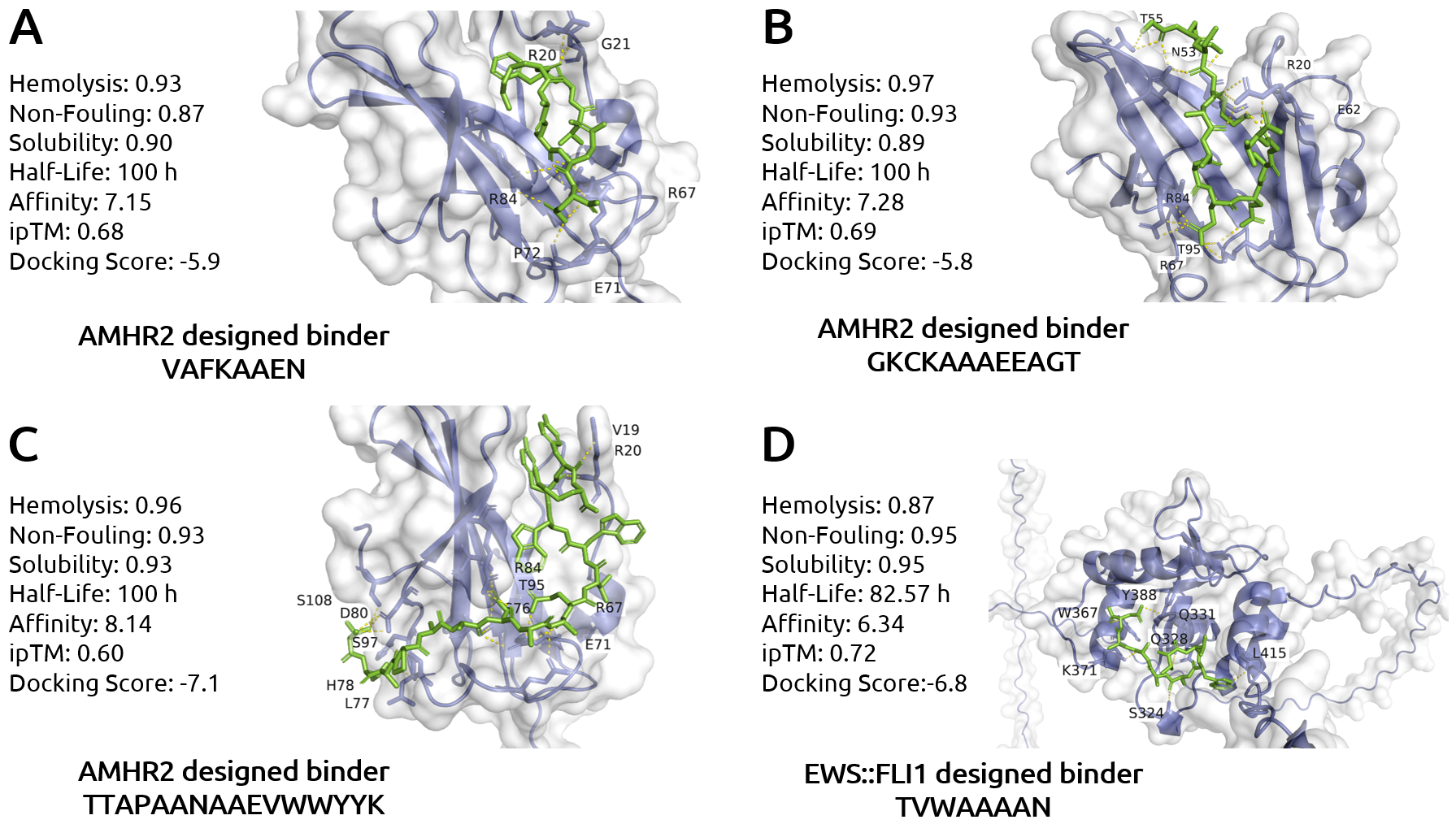}
    \par\vspace{0.7em}\par
    \includegraphics[width=\textwidth]{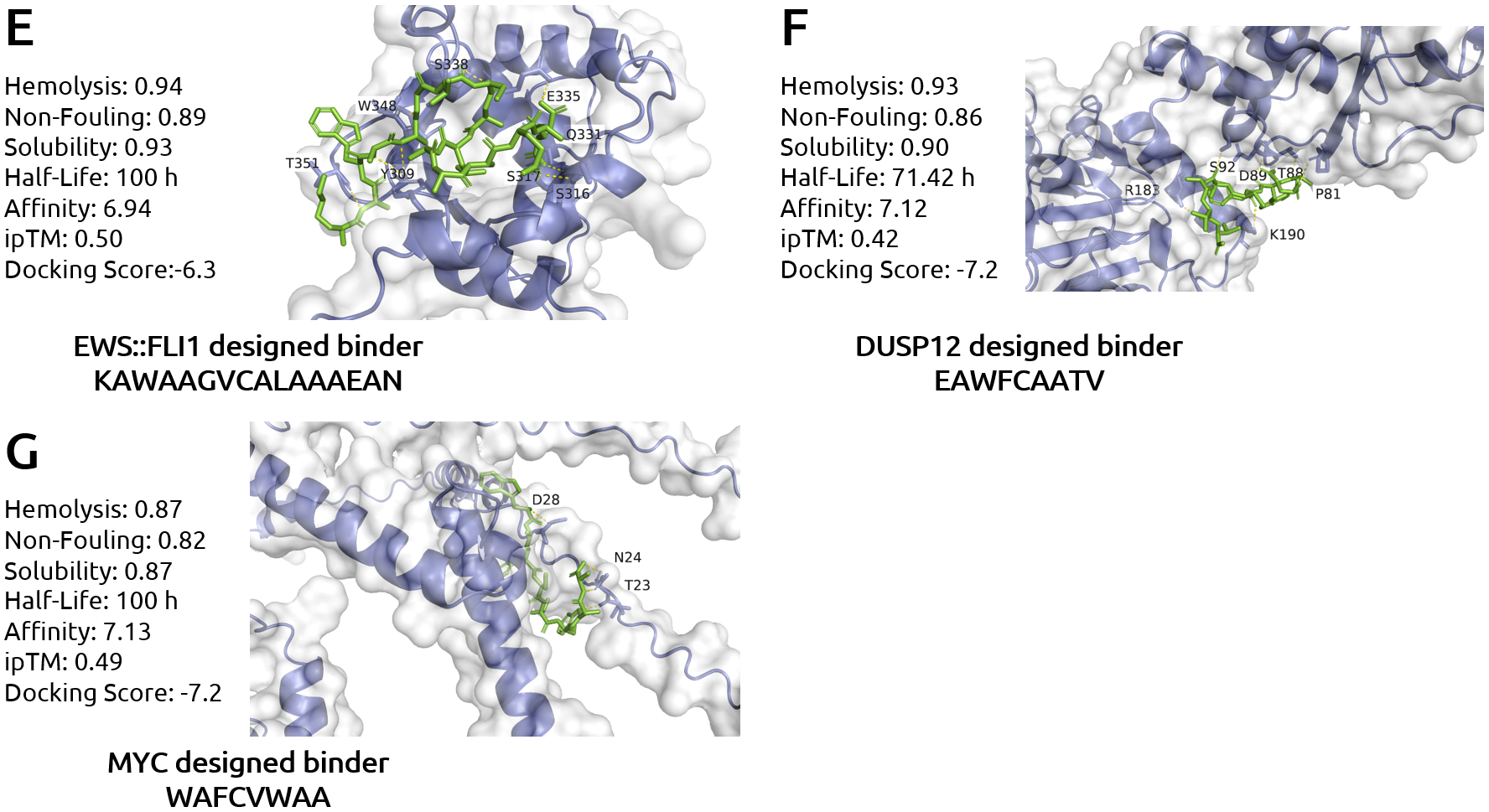}
    \caption{\textbf{Complex structures of target proteins without pre-existing binders.} \textbf{(A)-(C)} AMHR2, \textbf{(D)-(E)} EWS::FLI1, \textbf{(F)} MYC, \textbf{(G)} DUSP12. Each panel shows the complex structure of the target with an AReUReDi-designed binder. For each binder, five property scores are provided, as well as the ipTM score from AlphaFold3 and the docking score from AutoDock VINA. Interacting residues on the target are visualized.}
    \label{fig:w/o_binders}
\end{figure}
\begin{figure}[t]
    \centering
      \includegraphics[width=\textwidth]{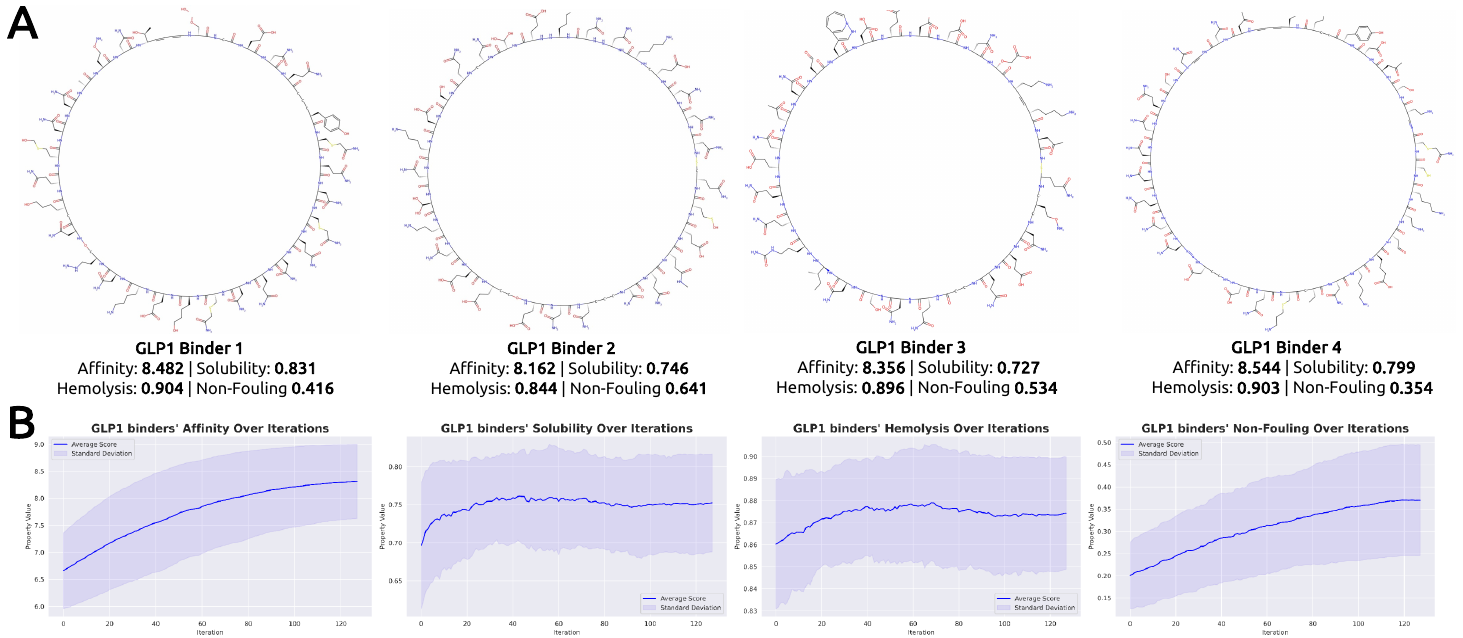}
      \vspace{2em}
      \includegraphics[width=\textwidth]{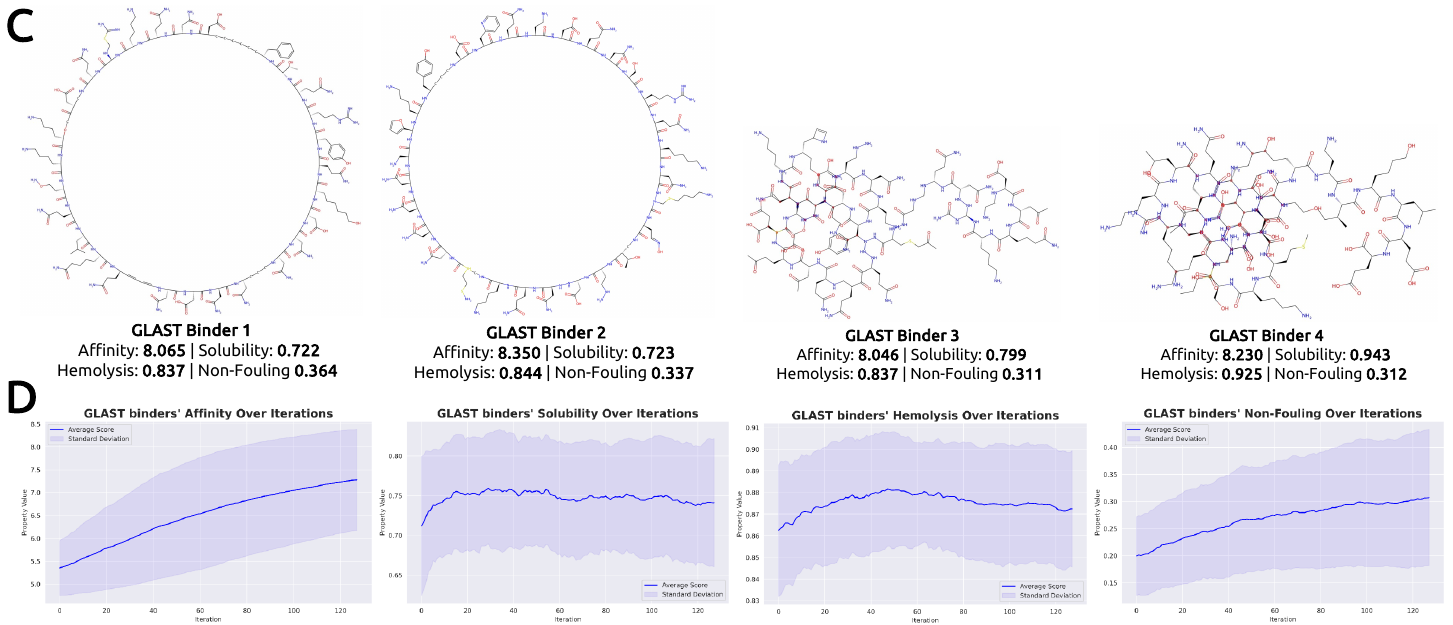}
      
    \caption{\textbf{(A), (C)} Example 2D SMILES structure of AReUReDi-designed peptide binders with four property scores for GLP1 and GLAST, respectively. \textbf{(B), (D)} Plots showing the mean scores for each property across the number of iterations during AReUReDi's design of binders of length 200 for GLP1 and GLAST, respectively.}
    \label{fig:GLP1_GLAST}
    \vskip -0.1in
\end{figure}
\begin{figure}[t]
    \centering
      \includegraphics[width=\textwidth]{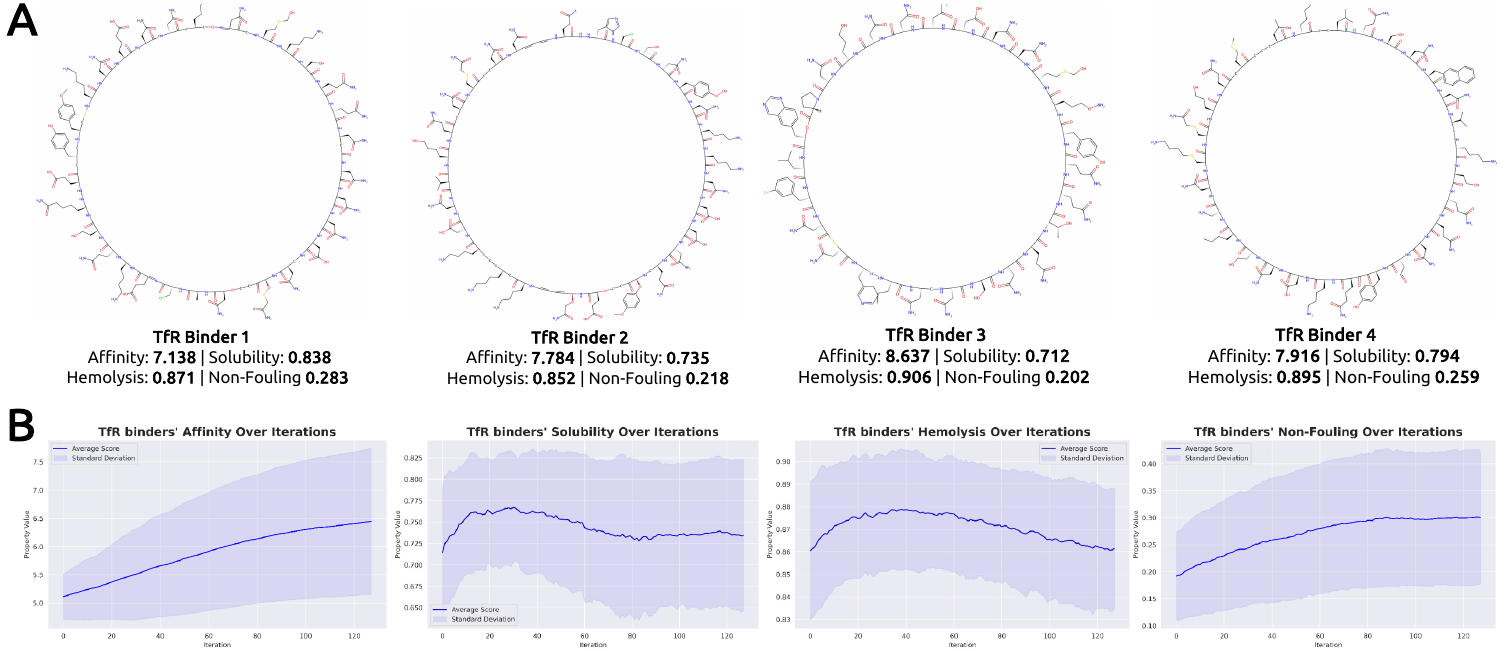}
      \vspace{2em}
      \includegraphics[width=\textwidth]{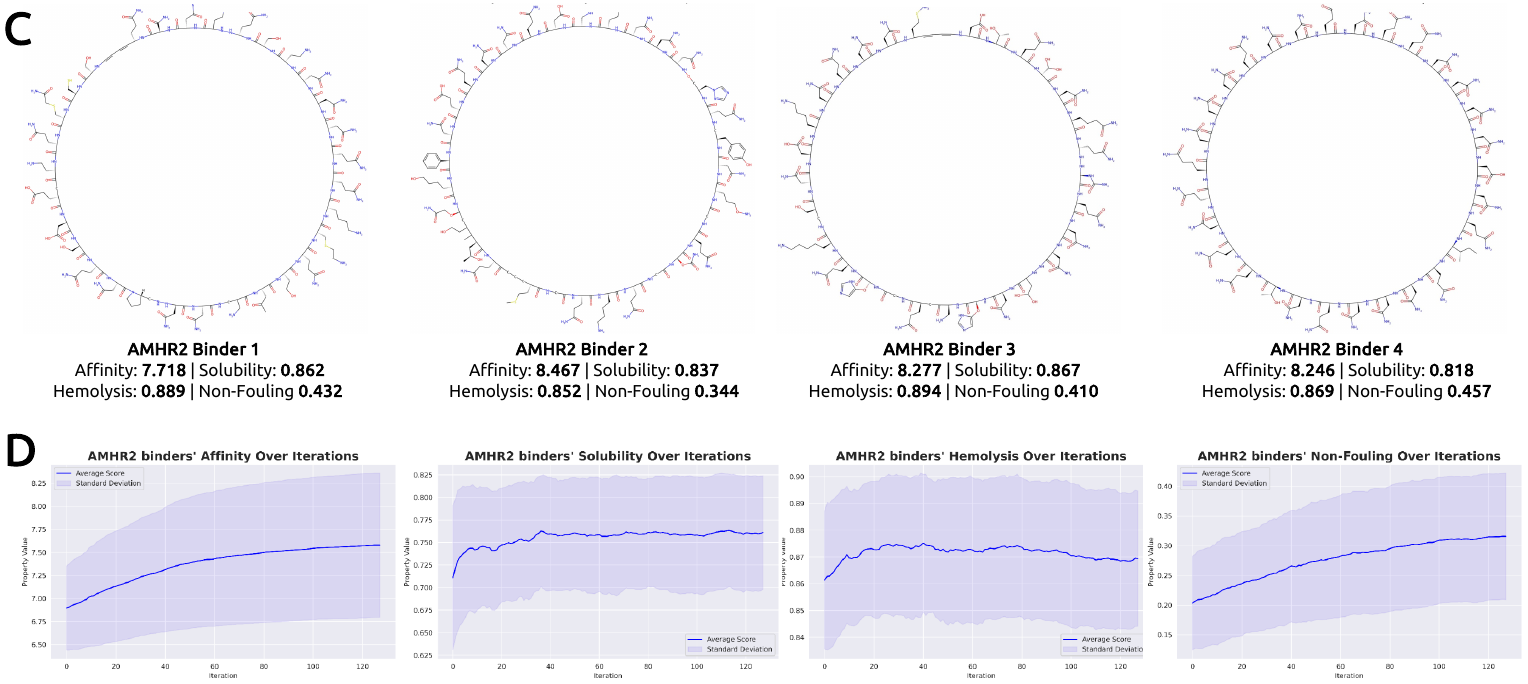}
      
    \caption{\textbf{(A), (C)} Example 2D SMILES structure of AReUReDi-designed peptide binders with four property scores for TfR and AMHR2, respectively. \textbf{(B), (D)} Plots showing the mean scores for each property across the number of iterations during AReUReDi's design of binders of length 200 for TfR and AMHR2, respectively.}
    \label{fig:TfR_AMHR2}
    \vskip -0.1in
\end{figure}
\begin{algorithm}
\caption{AReUReDi: Annealed Rectified Updates for Refining Discrete Flows}
\label{alg:areuredi}
\begin{algorithmic}[1]

\State \textbf{Input:} Pre-trained ReDi model $p_t^i(\cdot|x_t)$, objective functions $\tilde{s}_1, \dots, \tilde{s}_N$, weight vector $\omega \in \Delta^{N-1}$, annealing parameters $\eta_{min}, \eta_{max}$.
\State \textbf{Output:} Sequence $x_T$ with multi-objective optimized properties.
\State \strut

\State \textbf{Initialize:}
\State \quad Sample an initial sequence $x_0$ uniformly from the discrete state space $S$
\State \quad Sample or specify a weight vector $\omega \in \Delta^{N-1}$
\State \strut

\For{$t = 0$ to $1$ with step size $h = \frac{1}{T}$}

    % Step 1: Annealing and Coordinate Selection
    \State \textbf{Step 1: Annealing and Coordinate Selection}
    \State \quad Update guidance strength: $\eta_t \gets \eta_{min} + (\eta_{max} - \eta_{min})\frac{t}{T-1}$
    \State \quad Select a position $i$ in the sequence to update: $i \sim \text{Uniform}(\{1, \dots, L\})$
    \State \strut 

    % Step 2: Proposal Generation via Local Balancing
    \State \textbf{Step 2: Proposal Generation via Local Balancing}
    \State \quad Let $C_i$ be the set of candidate tokens from $p_t^i(\cdot|x_t)$.
    \State \quad For each candidate token $y \in C_i$:
        \State \quad\quad 1. Compute scalarized reward ratio:  $
        r_i(y; x_t) \gets \frac{\exp\left(\eta_t \min_{n} \omega_n \tilde{s}_n(x^{(i \leftarrow y)})\right)}{\exp\left(\eta_t \min_{n} \omega_n \tilde{s}_n(x)\right)}
        $
        \State \quad\quad 2. Compute unnormalized proposal distribution $\tilde{q}_i(y|x_t)$ using a balancing function $g(\cdot)$: $$\tilde{q}_i(y|x_t) \gets p_t^i(y|x_t) g(r_i(y; x_t))$$ \vspace{1em}
        \State \quad\quad 3. Normalize to get the final proposal distribution $q_i(y|x_t)$.
    \State \strut

    % Step 3: Metropolis-Hastings Acceptance
    \State \textbf{Step 3: Metropolis-Hastings Acceptance}
    \State \quad Sample a candidate token $y^* \sim q_i(\cdot|x_t)$.
    \State \quad Form the proposed state $x_{prop} \gets x^{(i \leftarrow y^*)}$.
    \State \quad Compute acceptance probability $\alpha_i(x, x_{prop})$: \[
    \alpha_i(x, x_{prop}) \gets \min\left\{1, \frac{\pi_{\eta_t, \omega}(x_{prop}) q_i(x^i|x_{prop})}{\pi_{\eta_t, \omega}(x) q_i(y^*|x)}\right\}, \quad \text{where } \pi_{\eta_t, \omega}(z) \propto p_1(z) \exp\left(\eta_t \min_{n} \omega_n \tilde{s}_n(z)\right)
    \]
    \vspace{1em}
    \State \quad With probability $\alpha_i(x, x_{prop})$, accept the proposal: $x \gets x_{prop}$.
    \State \quad Update time: $t \to t + h$
\EndFor
\State \textbf{Return:} Final sequence $x_1$.
\end{algorithmic}
\label{algo:areuredi}
\end{algorithm}
\end{document}